\documentclass{article}
\usepackage[utf8]{inputenc}
\usepackage{main}
\usepackage{microtype}
\usepackage{enumerate}
\usepackage[inline]{enumitem}
\usepackage{graphicx}
\usepackage{subfigure}
\usepackage{times}
\usepackage{latexsym}
\usepackage{amsmath}
\usepackage{float}
\usepackage{footnote}
\usepackage{enumitem}
\usepackage{bm}
\usepackage{arydshln}
\usepackage{booktabs}
\usepackage{multicol}
\usepackage{multirow}
\usepackage{color}
\usepackage{xcolor}     
\usepackage{colortbl}
\usepackage{bbding}
\usepackage{makecell}
\usepackage{mathtools}
\usepackage{imakeidx}
\usepackage{longtable}
\usepackage{wrapfig}
\makeindex
\usepackage{arydshln}
\usepackage{lipsum}
\usepackage{natbib}
\usepackage[toc]{multitoc}
\usepackage[edges]{forest}
\usepackage[normalem]{ulem}
\definecolor{mydarkblue}{rgb}{0,0.08,0.45}
\usepackage[colorlinks=true,linkcolor=mydarkblue,citecolor=mydarkblue,filecolor=mydarkblue,urlcolor=mydarkblue]{hyperref}
\usepackage{caption}
\usepackage{CJKutf8}
\usepackage{awesomebox} % for infobox
\usepackage{bbding}
\usepackage[most]{tcolorbox}
\usepackage{xcolor,soul}         % colors
\usepackage{makecell}

\DeclareUnicodeCharacter{2212}{-}

\usepackage[tikz]{bclogo}
\usepackage[framemethod=tikz]{mdframed}
\definecolor{bgblue}{RGB}{245,243,253}
\definecolor{ttblue}{RGB}{91,194,224}

\mdfdefinestyle{mystyle}{%
  rightline=true,
  innerleftmargin=10,
  innerrightmargin=10,
  outerlinewidth=3pt,
  topline=false,
  rightline=true,
  bottomline=false,
  skipabove=\topsep,
  skipbelow=\topsep
}

\newtcolorbox{myboxi}[1][]{
  breakable,
  title=#1,
  colback=red!5,
  colbacktitle=red!5,
  coltitle=black,
  fonttitle=\bfseries,
  bottomrule=0pt,
  toprule=0pt,
  leftrule=2pt,
  rightrule=2pt,
  titlerule=0pt,
  arc=0pt,
  outer arc=0pt,
  colframe=red,
}

\newtcolorbox{myboxnote}[1][]{
  breakable,
  title=#1,
  colback=orange!0,
  colbacktitle=orange!0,
  coltitle=black,
  fonttitle=\bfseries,
  bottomrule=0pt,
  toprule=0pt,
  leftrule=2pt,
  rightrule=2pt,
  titlerule=0pt,
  arc=0pt,
  outer arc=0pt,
  colframe=orange,
}

\newtcolorbox{myboxii}[1][]{
  breakable,
  freelance,
  title=#1,
  colback=white,
  colbacktitle=white,
  coltitle=black,
  fonttitle=\bfseries,
  bottomrule=0pt,
  boxrule=0pt,
  colframe=white,
  overlay unbroken and first={
  \draw[red!75!black,line width=3pt]
    ([xshift=5pt]frame.north west) -- 
    (frame.north west) -- 
    (frame.south west);
  \draw[red!75!black,line width=3pt]
    ([xshift=-5pt]frame.north east) -- 
    (frame.north east) -- 
    (frame.south east);
  },
  overlay unbroken app={
  \draw[red!75!black,line width=3pt,line cap=rect]
    (frame.south west) -- 
    ([xshift=5pt]frame.south west);
  \draw[red!75!black,line width=3pt,line cap=rect]
    (frame.south east) -- 
    ([xshift=-5pt]frame.south east);
  },
  overlay middle and last={
  \draw[red!75!black,line width=3pt]
    (frame.north west) -- 
    (frame.south west);
  \draw[red!75!black,line width=3pt]
    (frame.north east) -- 
    (frame.south east);
  },
  overlay last app={
  \draw[red!75!black,line width=3pt,line cap=rect]
    (frame.south west) --
    ([xshift=5pt]frame.south west);
  \draw[red!75!black,line width=3pt,line cap=rect]
    (frame.south east) --
    ([xshift=-5pt]frame.south east);
  },
}

\usepackage{fancyhdr} % to change header and footers
\usepackage{blindtext} % to quickly get a full document

\DeclareCaptionFont{black}{\color{black}}

\definecolor{myblue}{rgb}{0.9, 0.1, 0.94}
\definecolor{mygreen}{rgb}{0.64, 0.56, 0.88}
\definecolor{myyellow}{rgb}{0.68, 0.6, 0.1}
\definecolor{fancygreen}{rgb}{0.33, 0.68, 0.20}
\definecolor{salmon}{rgb}{0.94, 0.52, 0.49}
\definecolor{tablegreen}{rgb}{0.82, 0.94, 0.75}
\definecolor{tableblue}{rgb}{0.81, 0.90, 0.94}
\definecolor{tablered}{rgb}{0.97, 0.85, 0.85}
\definecolor{tableorange}{rgb}{0.96, 0.85, 0.81}

\usepackage{xcolor}
\usepackage{listings}

\newcommand\JSONnumbervaluestyle{\color{blue}}
\newcommand\JSONstringvaluestyle{\color{red}}

\newif\ifcolonfoundonthisline

\makeatletter

\lstdefinestyle{json}
{
  showstringspaces    = false,
  keywords            = {false,true},
  alsoletter          = 0123456789.,
  morestring          = [s]{"}{"},
  stringstyle         = \ifcolonfoundonthisline\JSONstringvaluestyle\fi,
  MoreSelectCharTable =%
    \lst@DefSaveDef{`:}\colon@json{\processColon@json},
  basicstyle          = \ttfamily,
  keywordstyle        = \ttfamily\bfseries,
}

\newcommand\processColon@json{%
  \colon@json%
  \ifnum\lst@mode=\lst@Pmode%
    \global\colonfoundonthislinetrue%
  \fi
}

\lst@AddToHook{Output}{%
  \ifcolonfoundonthisline%
    \ifnum\lst@mode=\lst@Pmode%
      \def\lst@thestyle{\JSONnumbervaluestyle}%
    \fi
  \fi
  \lsthk@DetectKeywords% 
}

\lst@AddToHook{EOL}%
  {\global\colonfoundonthislinefalse}

\makeatother

\usepackage{etoolbox}
\usepackage{natbib}
\usepackage{url}
\newcounter{bibcount}
\makeatletter
\patchcmd{\@lbibitem}{\item[}{\item[\hfil\stepcounter{bibcount}{[\thebibcount]}}{}{}
\renewcommand\NAT@bibsetup%
  [1]{\setlength{\leftmargin}{\bibhang}\setlength{\itemindent}{-\parindent}%
      \setlength{\itemsep}{\bibsep}\setlength{\parsep}{\z@}}
\makeatother

\definecolor{Blueback}{RGB}{218, 227, 243} 
\definecolor{Greenback}{RGB}{226, 240, 217}
\definecolor{Redback}{RGB}{251, 229, 214} 

\newcommand{\blueback}[1]{
  \begingroup
  \sethlcolor{Blueback}%背景色
  \textcolor{black}{\hl{#1}}%textcolor里面对应文字颜色
  \endgroup
}

\newcommand{\redback}[1]{
  \begingroup
  \sethlcolor{Redback}%背景色
  \textcolor{black}{\hl{#1}}%textcolor里面对应文字颜色
  \endgroup
}

\newcommand{\greenback}[1]{
  \begingroup
  \sethlcolor{Greenback}%背景色
  \textcolor{black}{\hl{#1}}%textcolor里面对应文字颜色
  \endgroup
}

\begin{document}
\newcommand{\modelname}{\textsc{ReAlign}\xspace}

\title{Reformatted Alignment} 

\author{Run-Ze Fan\textsuperscript{\rm{1,6}}\space\space \space\space \space\space\space
Xuefeng Li\textsuperscript{\rm{1,6}}\space\space \space\space \space\space\space
Haoyang Zou\textsuperscript{\rm{3,6}}\space\space \space\space \space\space\space
Junlong Li\textsuperscript{\rm{1,6}}\space\space \space\space \space\space\space
\textbf{Shwai He}\textsuperscript{\rm 4}\\
\textbf{Ethan Chern}\textsuperscript{\rm{1,6}}\space\space \space\space \space\space
\textbf{Jiewen Hu}\textsuperscript{\rm{5,6}}\space\space \space\space \space\space
\textbf{Pengfei Liu}\textsuperscript{\rm{1,2,6}}\thanks{~~Corresponding author}\\
    \textsuperscript{\rm 1}Shanghai Jiao Tong University\space\space
    \textsuperscript{\rm 2}Shanghai Artificial Intelligence Laboratory \\
    \textsuperscript{\rm 3}Fudan University \space\space
    \textsuperscript{\rm 4}University of Maryland, College Park \\
    \textsuperscript{\rm 5}CMU\space\space
    \textsuperscript{\rm 6}Generative AI Research Lab (GAIR)\\
    {\tt runze.fan@icloud.com},\space\space
 {\tt pengfei@sjtu.edu.cn}
}

\maketitle
\thispagestyle{fancy}
\fancyhead{}
\lhead{\includegraphics[height=0.67cm]{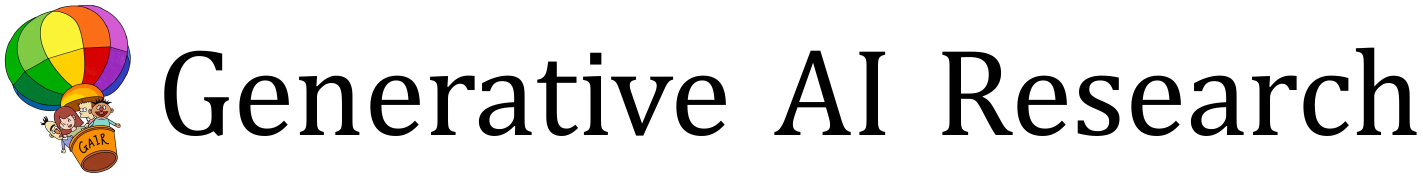}}
\renewcommand{\headrulewidth}{0pt}
\setlength{\headsep}{0mm}

\begin{abstract}
The quality of finetuning data is crucial for aligning large language models (LLMs) with human values.
Current methods to improve data quality are either labor-intensive or prone to factual errors caused by LLM hallucinations.
This paper explores elevating the quality of existing instruction data to better align with human values, introducing a simple and effective approach named \modelname, which \textit{reformats} the responses of instruction data into a format that better aligns with pre-established criteria and the collated evidence.
This approach minimizes human annotation, hallucination, and the difficulty in scaling, remaining orthogonal to existing alignment techniques.
Experimentally, \modelname significantly boosts the general alignment ability, math reasoning, factuality, and readability of the LLMs.

Encouragingly, \emph{without} introducing any additional data or advanced training techniques, and merely by reformatting the response, LLaMA-2-13B's mathematical reasoning ability on \texttt{GSM8K} can be improved \textbf{from 46.77\% to 56.63\%} in accuracy.
Additionally, a mere 5\% of \modelname data yields a 67\% boost in general alignment ability measured by the Alpaca dataset. 
This work highlights the need for further research into the \emph{science} and \emph{mechanistic interpretability} of LLMs. We have made the associated code and data publicly accessible to support future studies at 
\url{https://github.com/GAIR-NLP/ReAlign}.
\\ \\ \\ \\
\end{abstract}

\begin{figure}[h]
	\centering
	\subfigure[GSM8K]{
		\begin{minipage}[b]{0.48\columnwidth}
			\includegraphics[width=1\columnwidth]{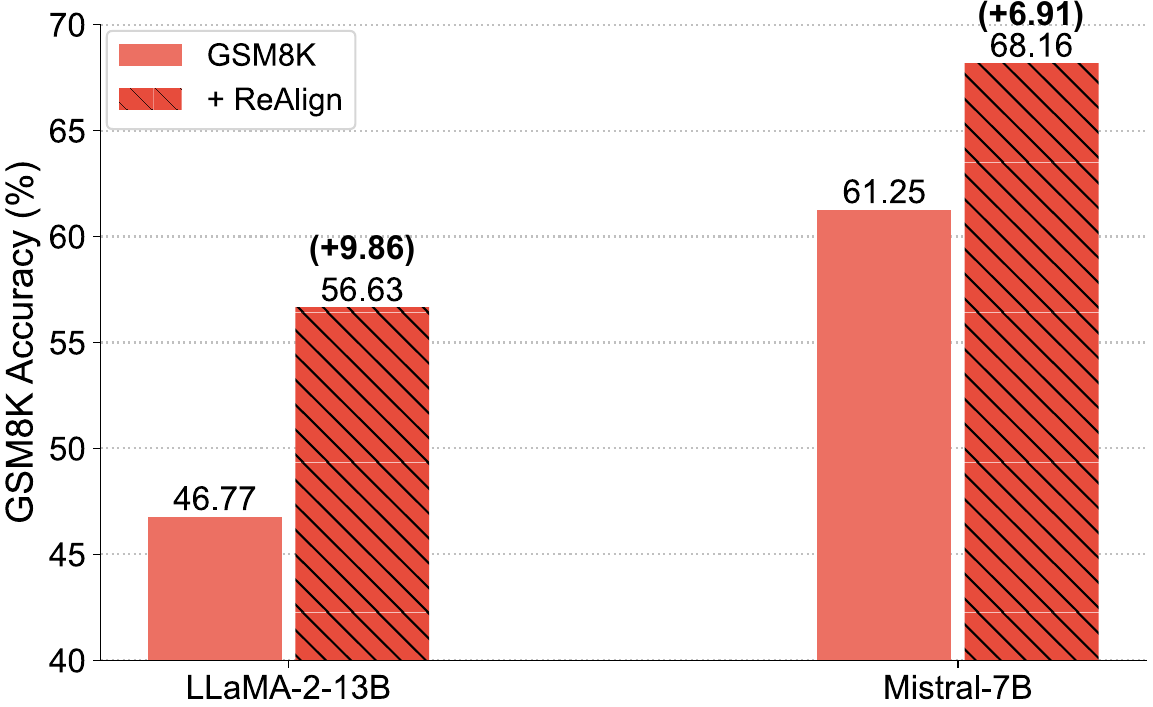} 
		\end{minipage}
	
	}
        % \hspace{20pt}
    	\subfigure[MATH (OOD Setting)]{
    		\begin{minipage}[b]{0.48\columnwidth}
   		 	\includegraphics[width=1\columnwidth]{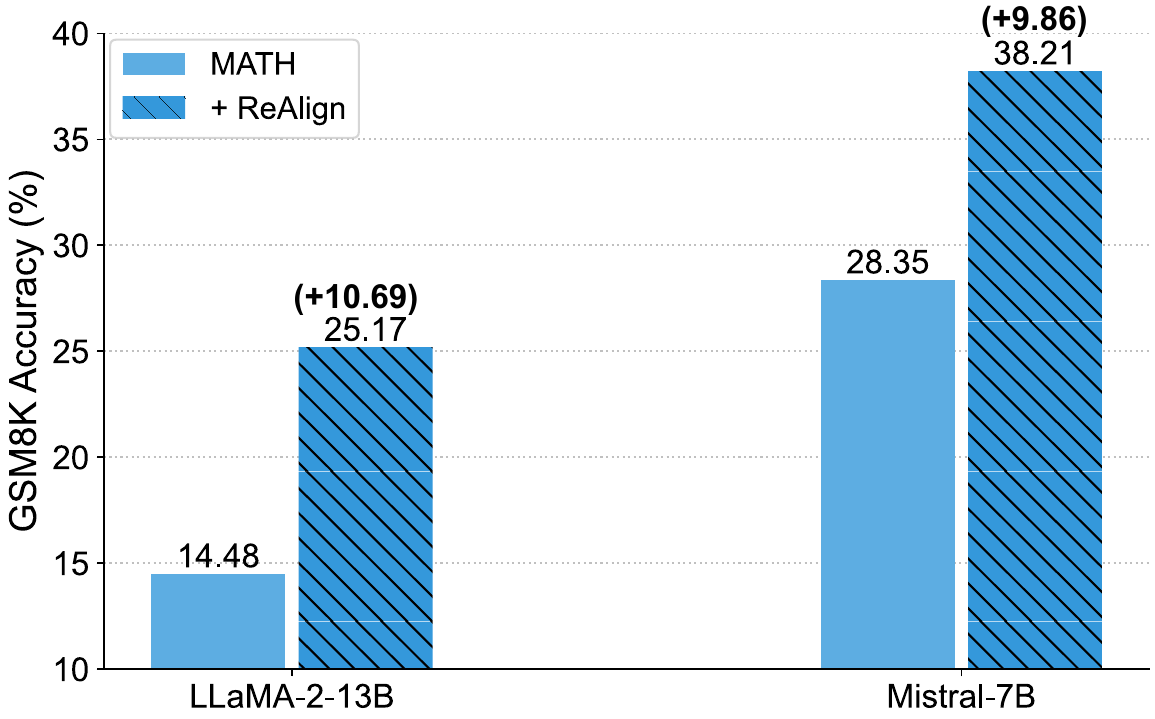}
    		\end{minipage}
	
    	}
     % \vspace{-8pt}
	\caption{The accuracy of the GSM8K test set for LLaMA-2-13B and Mistral-7B models fine-tuned on the training set of GSM8K and MATH with and without \modelname. (a): Training and testing on GSM8K. (b): Training on MATH and testing on GSM8K (Out-of-Distribution Setting).}
 % \vspace{-10pt}
\end{figure}

\newpage

\pagestyle{fancy}
\lhead{Reformatted Alignment}
% \chead{}
\renewcommand{\headrulewidth}{0.7pt}
\setlength{\headsep}{5mm}

% \tableofcontents

\clearpage

\section{Introduction}

Alignment has been witnessed to be an effective technique for aligning large language models (LLMs) to human values and human intent~\cite{ouyang2022rlhf}, which usually requires fine-tuning on a large amount of synthetic data derived from LLMs~\cite{wang-etal-2023-self-instruct,honovich-etal-2023-unnatural,peng2023instruction,xu2023wizardlm} or human-annotated instruction data~\cite{ouyang2022rlhf, köpf2023openassistant}.

Recent studies, notably by~\citet{zhou2023lima} highlight the critical role of instruction data quality in this process.
Numerous works~\cite{wang-etal-2023-self-instruct,zhou2023lima,cao2023instruction,chen2023alpagasus,li2023quantity,lu2023instag} have contributed to enhancing instruction quality by focusing on the diversity and complexity of input queries as well as the quality of responses. These efforts can be divided into two primary approaches.
The first approach, advocated by \citet{ouyang2022rlhf} and \citet{touvron2023llama2}, involves the manual creation of high-quality data. Although this method creates complex queries and factually correct and highly readable responses, it is labor-intensive and challenging to scale.
The second approach revolves around the automated extraction of high-quality instructions from existing datasets due to their extensive availability~\cite{cao2023instruction,chen2023alpagasus,li2023quantity,lu2023instag}. 
However, this method inherits the limitations associated with distilled data, such as containing factually incorrect content~\citep{ji-survey-hallucination-2023,gudibande2023false} and the format and style of the generated response are often determined by distilled LLMs' preference.

\begin{figure}[t]
\centering
\makeatother\def\@captype{figure}\makeatother
	\centering
	\includegraphics[width=0.8\columnwidth]{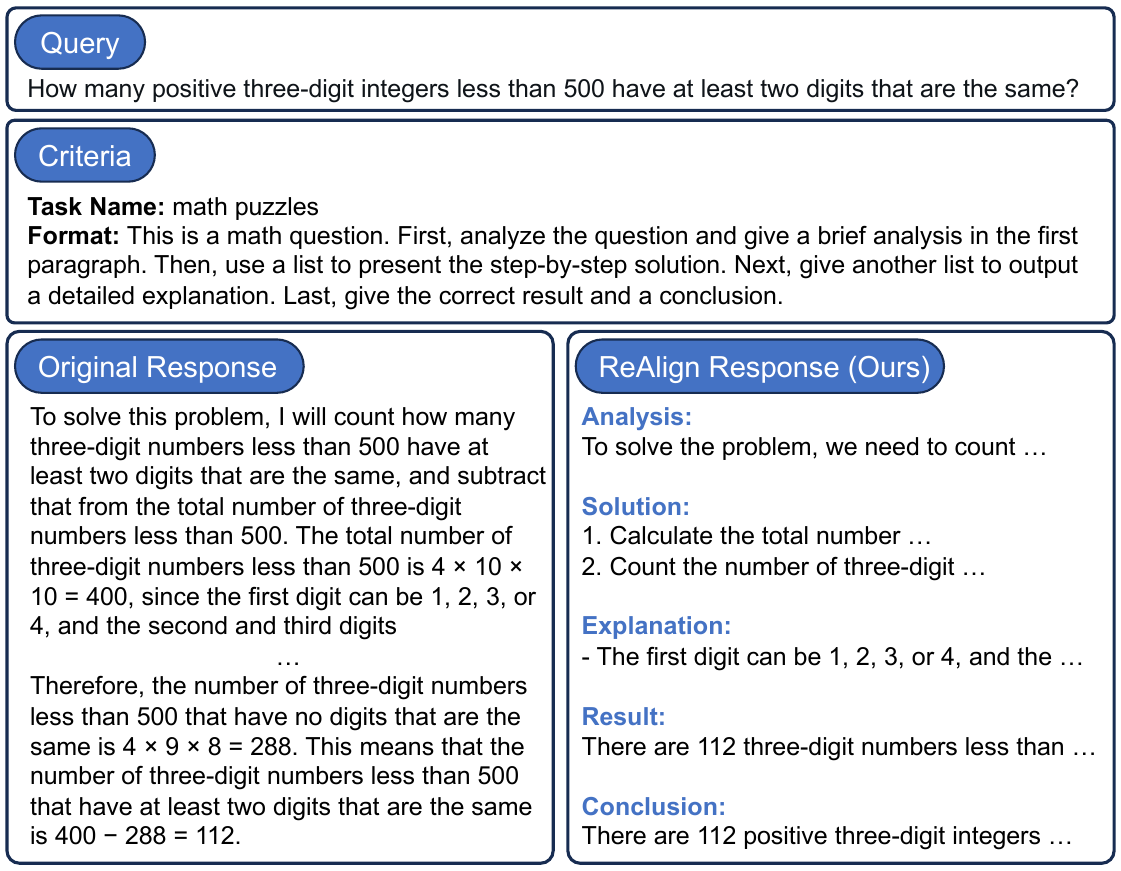}
 % \vspace{-20pt}
    \caption{\modelname realigns the original response with the pre-defined criteria to be a better format. The original response is from the \texttt{Open-Platypus}~\cite{lee2023platypus} dataset. The complete version is shown in Tab.~\ref{tab:complete_example}.}
    % \vspace{-20pt}
    \label{fig:example}
\end{figure}

In this paper, instead of focusing on the creation of instruction data from scratch, we investigate how existing instruction data can be made higher quality and better aligned with human values.
We propose a simple and effective method, named \modelname, which is orthogonal to the above existing approaches.
Specifically, \modelname necessitates a base instruction dataset, which can be sourced from extensive existing supervised datasets (e.g., \texttt{GSM8K}~\cite{cobbe2021gsm8k}), or publicly available instruction data compiled through various methods (e.g., Self-Instruct~\cite{wang-etal-2023-self-instruct}, Evol-Instruct~\cite{xu2023wizardlm}, and Self-Alignment~\cite{li2023selfalignment}). 
The \modelname process unfolds in three main steps. 
The first step involves \textbf{criteria definition}~(\S\ref{sec:mothed_tasks}), where humans define their preferences (e.g., the preferred format of responses) in various scenarios in the form of natural language. 
In this paper, we meticulously define criteria for 46 distinct scenarios. 
The second step, \textbf{retrieval augmentation}~(\S\ref{sec:rag}), broadens the knowledge base for knowledge-intensive tasks like open-domain QA and fact verification. This is achieved by incorporating additional information, thereby improving the factuality and informativeness of responses. 
The final step, \textbf{reformatting}~(\S\ref{sec:reformatting}), aims to re-align the responses with the pre-established criteria and the collated evidence, guaranteeing outputs that are both structured and substantiated.
As demonstrated in Fig.~\ref{fig:example}, the realigned response provides a better format and a clearer chain of thoughts.

The underlying \emph{philosophy} of \modelname is to re-coordinate the roles of humans and LLMs in the alignment process, leveraging their complementary strengths -- humans articulate their preferences, and LLMs, in turn, reconstruct instructions based on their generative power (e.g., instruction-following ability), without directly using distilled LLM knowledge.
Through this collaborative synergy, we expect the generated instruction data to be not only more contextually precise but also more closely aligned with human preferences.

We operationalize this idea on five types of existing instruction data, where three are general datasets (i.e., \texttt{Open-Platypus}~\cite{lee2023platypus}, \texttt{No Robots}~\cite{no_robots}, and \texttt{Alpaca}~\cite{alpaca}) and two are mathematical datasets (i.e., \texttt{GSM8K}~\cite{cobbe2021gsm8k} and \texttt{MATH}~\cite{hendrycksmath2021}).
The performance of \modelname has been validated across various well-established benchmarks, including AlpacaEval~\cite{alpaca_eval}, MT-Bench~\cite{zheng2023judging}, and Vicuna-Bench~\cite{vicuna2023} for general alignment, as well as \texttt{GSM8K} and \texttt{MATH} for mathematical reasoning. Additionally, it has also been evaluated for factuality and readability, demonstrating its proficiency.
In particular, \modelname significantly boosts math reasoning, even up to 9.86\% on \texttt{GSM8K} for LLaMA-2-13B.
Notably, we find that only 5\% of the \modelname data yields a 67\% boost in general alignment ability compared to the full \modelname data based on the \texttt{Alpaca} dataset, indicating that only a small amount of data is required to learn style and format.

\section{Related Work}

\subsection{Instruction Creation}

Creating instructional data significantly improves LLMs' alignment abilities.
High-quality instruction generation traditionally depends on human annotation for tasks like query writing, response drafting, and preference indication. This approach produces premium open-source datasets (e.g., \texttt{Open-Platypus}~\cite{lee2023platypus} and \texttt{OpenAssistant}~\cite{köpf2023openassistant}) and supports advanced LLMs (e.g., LIMA~\cite{zhou2023lima} and LLaMA-2~\cite{touvron2023llama2}), but it's hard to scale due to high labor costs and the need for domain-specific expertise.
Many studies have explored using LLMs (e.g., GPT-3~\cite{brown2020gpt3} and GPT-4~\cite{openai2023gpt4}) to generate instruction data. Techniques like unnatural instructions~\cite{honovich-etal-2023-unnatural} and self-instruct~\cite{wang-etal-2023-self-instruct} utilize GPT-3's in-context learning with seed data to generate instructions, while evol-instruct~\cite{xu2023wizardlm} generates more complex and varied instructions through ChatGPT. Recently, training with self-generated data has yielded excellent results, achieving self-alignment~\cite{li2023selfalignment,yuan2024selfrewarding,chen2024self}. While it can be easily scaled up, this approach inherits the drawbacks of LLMs (e.g., factual errors)~\cite{gudibande2023false}.
Our proposed method stands out by providing a way to automatically improve data quality with minimal effort and a significant reduction in factual errors.

\subsection{Instruction Selection}
After the discovery of ``quality is all you need''~\cite{zhou2023lima,touvron2023llama2}, instruction selection has been paid attention to, aiming at selecting a small number of the highest-quality samples from a large amount of instruction data as a training dataset.
\citet{cao2023instruction} evaluates the dataset's quality by utilizing the evaluation dataset loss to fit the natural language indicators of the dataset. 
\citet{chen2023alpagasus} proposes to use ChatGPT directly to score the data, while \citet{li2023selfalignment} proposes to score the data using the trained model directly to save costs.
\citet{lu2023instag} proposes to tag samples within SFT datasets based on semantics and intentions and define instruction diversity and complexity regarding tags to rank data.
\citet{li2023quantity} introduces a self-guided approach that utilizes a new indicator, Instruction-Following Difficulty (IDF), to score data by identifying gaps in a model's responses versus its autonomous generation capability.
\citet{anonymous2023what} trains two scorers to evaluate the complexity of the instruction and the quality of the response, respectively, and then uses the embedding distance to determine the diversity to select high-quality data.
However, the above works usually mine from distilled datasets because the large scale of distilled datasets is available, thereby inheriting the drawbacks of distilled data and suffering from the hallucination of LLMs.

\subsection{Instruction Tuning}
Instruction tuning aims to reinforce the model's instruction-following capabilities and align LLMs to human values.
Early instruction tuning was designed to improve cross-task generalization capabilities, in which they usually scale up the quantity and the diversity of tasks~\cite{mishra-etal-2022-cross,wei2022finetuned,sanh2022multitask,wang2022supernaturalinstructions}. Recent works no longer explicitly define tasks, but extend to more generalized capabilities, especially for scenarios of real-world questions~\cite{wang-etal-2023-self-instruct,honovich-etal-2023-unnatural,peng2023instruction,xu2023wizardlm}.
Differently, our work utilizes the future of the task to design a better format for it, which further improves the quality of the data.

\begin{table}[!ht]
  \centering
  \scriptsize
    \begin{tabular}{ll}
    \toprule
    \textbf{Group} & \textbf{Tasks} \\
    \midrule
    Generation & \makecell[l]{question generation; story generation; poem generation;  email generation; data generation; text-to-text translation}\\
    \midrule
    Brainstorming & \makecell[l]{advice giving; recommendations; how-to generation; planning}\\
    \midrule
    Code & \makecell[l]{code correction; code simplification; explain code;  text-to-code translation; code-to-code translation;\\ language learning questions;  code language classification;  code-to-text-translation}\\
    \midrule
    Rewriting & \makecell[l]{instructional rewriting; language polishing; paraphrasing;  text correction}\\
    \midrule
    Extraction & \makecell[l]{information extraction; keywords extraction; table extraction}\\
    \midrule
    Summarization & title generation; text summarization; note summarization\\
    \midrule
    Conversation & \makecell[l]{open qa; closed qa; fact verification; value judgment; roleplay; explain answer}\\
    \midrule
    Education. & \makecell[l]{natural language tutor; exam problem tutor; ai tutor; math puzzles; fill in the blank}\\
    \midrule
    Classification & \makecell[l]{general classification; ordering; sentiment analysis; language classification; topic classification}\\
    \midrule
    Others & rejecting; others\\
    \bottomrule
    \end{tabular}%
    % \vspace{-5pt}
  \caption{The category of tasks. ``{Education.}'' denotes {Specialized Educational Dialog.}}
  % \vspace{-10pt}
  \label{tab:tasks_category}
\end{table}

\begin{table}[!t]
    \scriptsize
    \centering
\begin{tabular}{@{}p{0.85\columnwidth}@{}}
\toprule
\textbf{Email Generation}\\

\makecell[l]{It is an email-writing task. Here is a general guideline for creating a well-structured and professional email:}

\\

\makecell[l]{\blueback{1. Subject Line:} Write a clear and concise subject line that accurately summarizes the content of your email …}

\makecell[l]{\blueback{2. Salutation:} Begin your email with a formal salutation such as "Dear [Recipient's Name]," …}

\blueback{3. Introduction:} Start your email with a brief introduction …

\blueback{4. Body:} This is the main content of your email …

\makecell[l]{\blueback{5. Politeness and Tone:} Maintain a polite and respectful tone throughout your email …}

\makecell[l]{\blueback{6. Closing:} Conclude your email with a closing remark, such as "Thank you," or "Best regards," followed by your name …}

\makecell[l]{\blueback{7. Signature:} Include your full name, job title, and contact information (e.g., phone number, email address) …}

\blueback{8. Attachments:} If you need to include attachments, mention them …

\makecell[l]{\blueback{9. Proofread:} Before sending the email, proofread it for any grammatical or spelling errors …}

\\

\makecell[l]{The best emails are short, direct, professional, and scannable for the recipient. Follow a formal business email structure unless you have \\ an established casual rapport with the recipient.}
\\
\bottomrule
\end{tabular}
% \vspace{-5pt}
    \caption{An example of the format for the ``email generation'' task.}
    \label{tab:format_example}
    % \vspace{-18pt}
\end{table}

\begin{figure*}[t]
\centering
\makeatother\def\@captype{figure}\makeatother
	\centering
	\includegraphics[width=0.9\columnwidth]{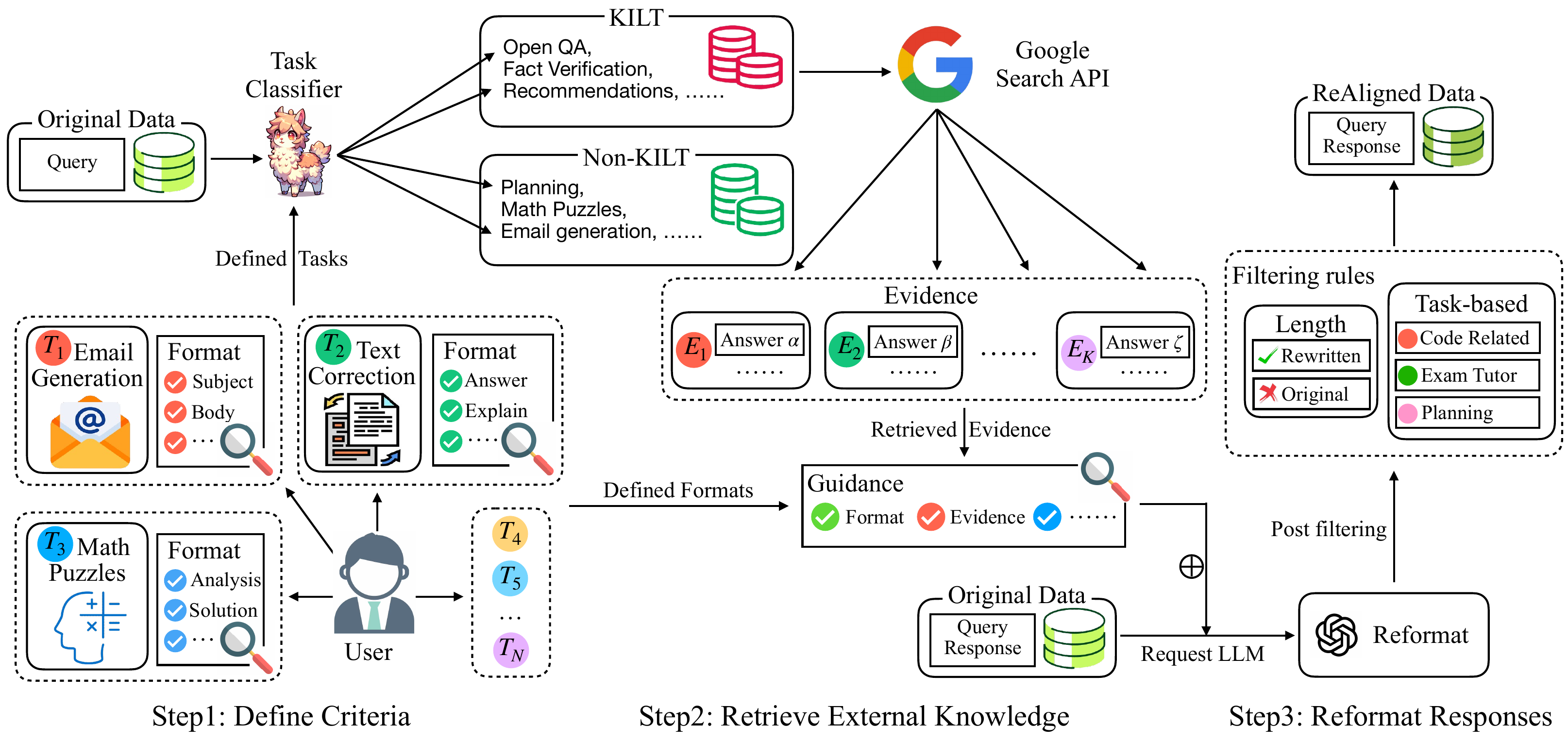}
 % \vspace{-5pt}
    \caption{An overview of our \modelname including three steps. KILT denotes Knowledge Intensive Language Tasks.}
    % \vspace{-12pt}
    \label{fig:overview}
\end{figure*}

% \vspace{-5pt}
\section{\modelname}
% \vspace{-10pt}
\label{sec:method}
Given a base instruction dataset $\mathcal{D} = \{(q_1, r_1),\cdots, (q_n, r_n)\}$, where $q$ and $r$ are the input query and response respectively, \modelname aims to improve the quality of responses by three steps as shown in Fig.~\ref{fig:overview}: 
(1) Criteria Definition: defining the criteria including tasks and formats for each task, (2) Retrieval Augmentation: retrieving relevant external information for the knowledge-intensive tasks, and (3) Reformatting: reformatting the original response based on the guidance consisting of hand-written format and the retrieved information. An overview of our method is shown in Fig.~\ref{fig:overview}.

% \vspace{-5pt}
\subsection{Criteria Definition}
% \vspace{-5pt}
\label{sec:mothed_tasks}
The predefined criteria consist of the tasks and the corresponding formats:

\paragraph{Tasks.}
Clearly defining tasks is crucial to subsequently devising tailored formats, as the optimal format varies across distinct tasks.
In this paper, we follow \citet{li2023generative} to define 46 different tasks $\{ T_1, \cdots, T_{N=46} \}$, categorized into 10 major groups, as shown in Tab.~\ref{tab:tasks_category}. The detailed description for each task is shown in Tab.~\ref{tab:tasks_description}, \S\ref{sec:tasks_description}. We also train a task classifier $\mathtt{C}$, detailed in \S\ref{sec:task_classifier}.
% \vspace{-3pt}
\paragraph{Format.}
Due to the distinct formatting requisites associated with diverse tasks, we meticulously devised tailored formats $\{ F_1, \cdots, F_{N=46}\}$ for each task based on the task definition and description, encompassing considerations such as organizational structure, section content, and output modality.
Specifically, we summarize the formatting deficiencies of GPT-4, Claude-2, and Bard in each task, and investigate user preferences for formatting regarding these tasks. Based on this information, we design tailored formats, which are more readable than the generic format.
% This format, which has been carefully designed for different tasks, is more readable than the generic format.
Each format has a task name and a detailed format description. We show an example of a format for ``email generation'' in Tab.~\ref{tab:format_example} (The complete version is shown in Tab.~\ref{tab:complete_format}).

In this step, we input query $q_i$ to the task classifier $\mathtt{C}$ (detailed in \S\ref{sec:task_classifier}) to acquire the category $t_i$:
% \vspace{-4pt}
\begin{equation}
% \vspace{-3pt}
     t_i = \mathtt{C}(q_i),
     \nonumber
     % \vspace{-3pt}
\end{equation}
% \vspace{-4pt}
 and then obtain the corresponding format $f_i$.

% \vspace{-4pt}
\subsection{Retrieval Augmentation}
% \vspace{-3pt}
\label{sec:rag}
Knowledge-intensive language tasks (KILT), such as open-domain QA and fact verification, usually require large and external knowledge sources as the evidence to ensure the factuality~\cite{petroni-etal-2021-kilt}.
Thus, we follow \citet{petroni-etal-2021-kilt} to choose five knowledge-intensive tasks and use the query $q_i$ to retrieve relevant information as our evidence. The tasks for retrieval augmentation are shown in Tab.~\ref{tab:tasks_description}. Specifically, we follow~\citet{chern2023factool} and use the Google Search API as our retriever $\mathtt{R}$ provided by Serper\footnote{\url{https://serper.dev/}} to retrieve the most relevant search snippets included in the API’s answer. We then parse the response to obtain different types of snippets such as answer boxes, knowledge graphs, and organic search results. Finally, we choose the top-$k$ snippets and filter them as our evidence $E_i={e_{i1},\cdots,e_{ik}}$:
% \vspace{-3pt}
\begin{equation}
% \vspace{-4pt}
     E_i = \mathtt{R}(q_i).
     \nonumber
     % \vspace{-4pt}
\end{equation}
We show an example of a knowledge-intensive language task in Tab.~\ref{tab:kilt_example}, demonstrating that retrieval augmentation enables the response more factual and informative.

% \vspace{-3pt}
\subsection{Reformatting}
\label{sec:reformatting}
% \vspace{-3pt}
\subsubsection{Rewriting}
% \vspace{-3pt}
In this step, we leverage large language models (e.g., ChatGPT) to rewrite the response $r_i$ based on the given format $f_i$ and retrieved evidence $E_i$ (for knowledge-intensive tasks). 
Since certain queries have additional requirements (e.g., specific formatting or specified information), an adaptive rewriting strategy is employed.
This approach involves initially using LLMs to determine whether the format matches the query requirements.
Subsequently, if it matches, the LLMs rewrite the response accordingly.
We divide the tasks into two categories:
% \vspace{-3pt}
\paragraph{Non-knowledge-intensive tasks}
% \vspace{-3pt}
For the non-knowledge-intensive tasks, we decide to rewrite a part of the tasks. This decision stems from the observation that certain tasks are not amenable to a standardized format, exemplified by instances such as story generation and poem generation (see Tab.~\ref{tab:tasks_description} for details). We guide LLMs to rewrite the original responses $r_i$, organizing the query $q_i$, original response $r_i$, and the format $f_i$ together via the prompt in Tab.~\ref{tab:rewriting_prompts_non_retrieval}:
% \vspace{-3pt}
\begin{equation}
% \vspace{-3pt}
     \hat{r}_i = \mathtt{LLM}(q_i, r_i, f_i),
     \nonumber
     % \vspace{-3pt}
\end{equation}
% \vspace{-3pt}
where $\hat{r}_i$ is the reformatted response.
% \vspace{-3pt}
\paragraph{Knowledge-intensive tasks.}
For the knowledge-intensive tasks, we additionally utilize the retrieved evidence $E_i$ compared to non-knowledge-intensive tasks. Specifically, We guide LLM to rewrite the original response $r_i$, organizing the query $q_i$, original response $r_i$, format $f_i$, and the retrieved evidence $E_i$ together via the prompt in Tab.~\ref{tab:rewriting_prompts_retrieval}:

\begin{equation}
% \vspace{-3pt}
     \hat{r}_i = \mathtt{LLM}(q_i, r_i, f_i, E_i).
     \nonumber
     % \vspace{-3pt}
\end{equation}

% \vspace{-3pt}
\subsubsection{Post-processing}
% \vspace{-3pt}
\paragraph{Length filtering.}
We find that LLMs sometimes fail to reformat and only output the changed sentences, whose output length plummets.
To filter out the data that fails to be reformatted, we keep the original response instead of using the reformatted response that is less than half the length of the original response.

% \vspace{-3pt}
\paragraph{Task-based filtering.}
To mitigate the problem of error propagation in task classification, we design filtering rules for specific tasks: 
\begin{enumerate*}[label=(\roman*)]
\item For code-related tasks (e.g., ``\texttt{code correction}''), the keyword matching rule is employed to ascertain whether both the original and the reformatted versions contain code. If only one of the original responses or the reformatted response incorporates code, it signifies a failure in reformatting, and the original response is retained.
\item For the ``\texttt{exam problem tutor}'' task, reformatted responses that do not contain the accurate result will not be accepted.
\item For the ``\texttt{planning}'' task, if the query does not contain a planning-related keyword (e.g., plan or planning), the original answer is retained.
\end{enumerate*}

Finally, we could acquire the reformatted dataset $\mathcal{\hat{D}} = \{(q_1, \hat{r}_1),\cdots, (q_n, \hat{r}_n)\}$ (denotes as \modelname dataset).

% \vspace{-5pt}
\section{Experiments}
% \vspace{-3pt}
\subsection{Datasets}
% \vspace{-3pt}
For evaluation of general ability, we select two high-quality manual datasets and one distillation dataset for instruction tuning:
(1) \textbf{\texttt{Open-Platypus}}~\cite{lee2023platypus} is an amalgamation of 11 open-source datasets, carefully curated to enhance LLM performance in STEM and logical domains. It consists of 25k questions, with around 90\% written by humans and the rest generated by LLM.
(2) \textbf{\texttt{No Robots}}~\cite{no_robots} is a high-quality dataset of 10k instructions and demonstrations created by skilled human annotators.
(3) \textbf{\texttt{Alpaca}}~\cite{alpaca} is an open-source instruction tuning dataset generated from text-davinci-003~\cite{ouyang2022rlhf} by the \textit{Self-Instruct}~\cite{wang-etal-2023-self-instruct} method, containing 52k samples.
Additionally, we also choose two manual datasets to evaluate the math reasoning after using \modelname:
(4) \textbf{\texttt{GSM8K}}~\cite{cobbe2021gsm8k} is a high-quality grade school math problems dataset created by human problem writers, consisting of 7.5k training problems and 1k test problems.
(5) \textbf{\texttt{MATH}}~\cite{hendrycksmath2021} is a dataset of mathematics competitions problems, including 7.5k for training and 5k for testing.

\subsection{Models}
% \vspace{-3pt}
We select two well-known open-source base models for fine-tuning:
(1) \textbf{LLaMA-2-13B}~\cite{touvron2023llama2} is a open-source pre-trained model using 2T tokens.
(2) \textbf{Mistral-7B}~\cite{jiang2023mistral} is the current state-of-the-art base language model at the 7B parameter scale.

% \vspace{-3pt}
\subsection{Evaluation}
% \vspace{-3pt}
We evaluate \modelname on general alignment and specific alignment ability including math reasoning, factuality, and readability.

% \vspace{-3pt}
\subsubsection{General Alignment}
% \vspace{-3pt}
To evaluate the general alignment ability, we follow~\citet{wang2023openchat} to employ the most widely recognized benchmarks, including: \textbf{AlpacaEval}~\cite{alpaca_eval}, \textbf{MT-Bench}~\cite{zheng2023judging}, \textbf{Vicuna-Bench}~\cite{vicuna2023}. Specifically, we use GPT-3.5 and Auto-J (detailed in \S\ref{sec:description_autoj}) as the evaluators for AlpacaEval due to the cost of GPT-4, which has an extremely strong correlation with human~\cite{li2023generative, sun2024metacritique}, and GPT-4 for MT-Bench and Vicuna-Bench.
% \vspace{-3pt}
\subsubsection{Specific Alignment}
% \vspace{-3pt}
We evaluate specific perspectives for alignment, including math reasoning, factuality, and readability.

\begin{table*}[t]
\small
  \centering
  % \scriptsize
  \setlength{\tabcolsep}{5pt}
    \begin{tabular}{clccccccc}
    \toprule
    \multirow{2}{*}{Model}&\multirow{2}{*}{Dataset} & \multicolumn{2}{c}{AlpacaEval} & \multicolumn{3}{c}{MT-Bench} & \multirow{2}{*}{Vicuna-Bench} & \multirow{2}{*}{Overall} \\
    
    \cmidrule(lr){3-4}\cmidrule(lr){5-7} & & \multicolumn{1}{c}{GPT-3.5 (\%)} & \multicolumn{1}{c}{Auto-J} & \multicolumn{1}{c}{First} & \multicolumn{1}{c}{Second} & \multicolumn{1}{c}{Average} & & \\
    % \midrule
    % \multicolumn{6}{c}{\textit{Open-Platypus}}\\
    \midrule
    \multirow{8}{*}{LLaMA-2-13B} & \texttt{Open-Platypus} & 55.71 & \textbf{4.93} & 6.69  & \textbf{5.16} & \textbf{5.94} & 8.28 & 6.18 \\
    & \ \ + \modelname &\textbf{58.20} & 4.81 & \textbf{6.89}  & 4.86  & 5.88  & \textbf{8.45} & \textbf{6.24}  \\
    \cmidrule(lr){2-9}
    % \multicolumn{6}{c}{\textit{No Robots}}\\
    % \midrule
    & \texttt{No Robots} & 44.25 & 4.56 & 5.80  & 5.15 & 5.48 & 7.31 & 5.44  \\
    & \ \ + \modelname &\textbf{48.13} & \textbf{4.65} & \textbf{6.04}  & \textbf{5.20} & \textbf{5.62} & \textbf{7.51} & \textbf{5.65}  \\
    \cmidrule(lr){2-9}
    % \multicolumn{6}{c}{\textit{Alpaca}}\\
    % \midrule
    & \texttt{Alpaca} & 46.08 & 4.65 & 5.55  & 4.16 & 4.86 & 6.55 & 5.17 \\
    & \ \ + \modelname &\textbf{49.19} & \textbf{4.74} & \textbf{5.83}  & \textbf{4.71} & \textbf{5.27}  & \textbf{6.84} & \textbf{5.44} \\
    \midrule
    % \multicolumn{6}{c}{\textit{Average}}\\
    % \midrule
    % & Average of Original Datasets & 48.68  & 6.01  & 4.82 & 5.43 & 7.38 \\
    % & Average + \modelname &\textbf{51.84}  & \textbf{6.25}  & \textbf{4.92} & \textbf{5.59}  & \textbf{7.60}  \\
    \multirow{8}{*}{Mistral-7B} & \texttt{Open-Platypus} & 59.63 & 5.15 & 7.29 & 5.88 & 6.58 & \textbf{8.96} & 6.66 \\
    & \ \ + \modelname  & \textbf{61.33} & \textbf{5.15} & \textbf{7.43} & \textbf{6.18} & \textbf{6.80} & 8.86 & \textbf{6.74} \\
    \cmidrule(lr){2-9} 
    & \texttt{No Robots} & 44.22 & 4.62 & 5.95 & \textbf{4.94} & 5.44 & 7.32 & 5.45 \\
    & \ \ + \modelname & \textbf{48.26} & \textbf{4.76} & \textbf{6.14} & 4.79 & \textbf{5.46} & \textbf{7.68} & \textbf{5.68 }\\
    \cmidrule(lr){2-9}
    & \texttt{Alpaca} & 51.24 & 4.77 & 6.06 & \textbf{5.26} & 5.66 & 7.14 & 5.67 \\
    & \ \ + \modelname & \textbf{52.67} & \textbf{4.82} & \textbf{6.50} & 5.03 & \textbf{5.76} & \textbf{7.33} & \textbf{5.79} \\
    % \cmidrule(lr){2-7}
    \bottomrule
    \end{tabular}%
    % \vspace{-5pt}
  \caption{The results of the general alignment ability on the original datasets and the \modelname datasets. \textbf{Bold} indicates the best result on each dataset. For AlpacaEval, GPT-3.5 denotes the winning rate obtained by using GPT-3.5 as the evaluator. Auto-J denotes the quality of the model's responses evaluated in a point-wise manner using Auto-J~\cite{li2023generative}. For MT-Bench, we report the result of the first turn, the second turn, and the average, respectively. For Overall, we calculate the average of AlpacaEval's winning rate for GPT-3.5 divided by 10, the results for Auto-J, the average MT-Bench results, and the results for Vicuna-Bench.} 
  % \vspace{-17pt}
  \label{tab:alignment_ability}
\end{table*}

% \vspace{-3pt}
\paragraph{Math Reasoning.}
To evaluate math reasoning, we finetune LLaMA-2-13B and Mistral-7B on \texttt{GSM8K} and \texttt{MATH} training datasets, respectively, and test afterward. The prompt template for training and testing is \texttt{``Question:\textbackslash n\\\{input\}\textbackslash n Answer:\textbackslash nLet's think step by step.\textbackslash n''}.
Since both datasets consist of math problems in the same style, we apply forced rewriting instead of adaptive, which does not require the determination of whether the query and format match but rather mandates a rewriting. We determine the accuracy by extracting the last number from the responses and comparing it directly to the ground truth.

% % \begin{table}[t]
% \begin{wraptable}[r][9cm]
%   \centering
%   % \scriptsize
%   \setlength{\tabcolsep}{5pt}
%     % \fontsize{9}{12}\selectfont
%     \begin{tabular}{c|c|cc|c}
%     \toprule
%     Model & Dataset & \texttt{GSM8K} & \texttt{MATH} & Overall \\
%     \midrule
%     \multirow{4}{*}{LLaMA-2-13B}    & \texttt{GSM8K}     & 46.77 & 5.02  & 25.90\\
%                                     & + \modelname  & \textbf{56.63} & \textbf{5.46} & \textbf{31.05}\\
%     \cmidrule{2-5}
%                                     & \texttt{MATH}      & 14.48 & 6.14 &   10.31\\
%                                     & + \modelname  & \textbf{25.17} & \textbf{7.14} & \textbf{16.16}\\
%     \cmidrule{1-5}
%     \multirow{4}{*}{Mistral-7B}     & \texttt{GSM8K}     & 57.62 & 7.68 & 32.65\\
%                                     & + \modelname  & \textbf{62.47} & \textbf{9.02} & \textbf{35.75}\\
%     \cmidrule{2-5}
%                                     & \texttt{MATH}      & 28.35 & 13.18 & 20.77\\
%                                     & + \modelname  & \textbf{38.21} & \textbf{15.30} & \textbf{26.76}\\
%     \bottomrule
%     \end{tabular}%
%     % \vspace{-5pt}
%   \caption{The results of math reasoning on \texttt{GSM8K}, \texttt{MATH} and them + \modelname based on LLaMA-2-13B and Mistral-7B. We test models on both \texttt{GSM8K} and \texttt{MATH} test sets. We report the accuracy by exact matching. \textbf{Bold} indicates the best result.}.
%   % \vspace{-22pt}
%   \label{tab:math_reasoning}
%   \end{wraptable}
% % \end{table}

% \vspace{-3pt}
\paragraph{Factuality.}
\label{sec:evaluation_factuality}
To evaluate the factuality, we randomly select 100 cases from the Natural Questions dataset (NQ)~\cite{nq}, a public Q\&A dataset rich in fact-based queries and their verified answers. 
We employ both GPT-4 and human evaluation in our experiment. Specifically, GPT-4 is used to rate these instances on a factuality scale of 1 to 10, considering the question, the response, and the ground truth (referred to as the factuality score). The evaluation prompt is shown in Tab.~\ref{tab:factuality_evaluation_prompt}. For the human evaluation, we compare the model's response to the ground truth based on the question, assigning a True or False. The results we present are the proportion of responses that were assigned as True.
\vspace{-3pt}

% \begin{table}[t]
\begin{wraptable}{r}{8.6cm}
  \centering
  % \scriptsize
  \setlength{\tabcolsep}{5pt}
    % \fontsize{9}{12}\selectfont
    \begin{tabular}{c|c|cc|c}
    \toprule
    Model & Dataset & \texttt{GSM8K} & \texttt{MATH} & Overall \\
    \midrule
    \multirow{4}{*}{LLaMA-2-13B}    & \texttt{GSM8K}     & 46.77 & 5.02  & 25.90\\
                                    & + \modelname  & \textbf{56.63} & \textbf{5.46} & \textbf{31.05}\\
    \cmidrule{2-5}
                                    & \texttt{MATH}      & 14.48 & 6.14 &   10.31\\
                                    & + \modelname  & \textbf{25.17} & \textbf{7.14} & \textbf{16.16}\\
    \cmidrule{1-5}
    \multirow{4}{*}{Mistral-7B}     & \texttt{GSM8K}     & 57.62 & 7.68 & 32.65\\
                                    & + \modelname  & \textbf{62.47} & \textbf{9.02} & \textbf{35.75}\\
    \cmidrule{2-5}
                                    & \texttt{MATH}      & 28.35 & 13.18 & 20.77\\
                                    & + \modelname  & \textbf{38.21} & \textbf{15.30} & \textbf{26.76}\\
    \bottomrule
    \end{tabular}%
    % \vspace{-5pt}
  \caption{The results of math reasoning on \texttt{GSM8K}, \texttt{MATH} and them + \modelname based on LLaMA-2-13B and Mistral-7B. We test models on both \texttt{GSM8K} and \texttt{MATH} test sets. We report the accuracy by exact matching. \textbf{Bold} indicates the best result.}.
  % \vspace{-22pt}
  \label{tab:math_reasoning}
  \end{wraptable}
% \end{table}

\paragraph{Readability.}
To evaluate the readability, we compare a model trained on the original dataset against another model on the dataset enhanced with \modelname, using human and GPT-4 evaluations on the Vicuna-Bench dataset~\cite{vicuna2023}. Since the vicuna bench contains fewer complex questions (e.g., code and math), the judge can focus on the format rather than the result. We design an evaluation prompt prioritizing readability, refer to Tab.~\ref{tab:readability_evaluation_prompt}, and randomize response positions to eliminate bias.

% \vspace{-5pt}
\subsection{Results}
% \vspace{-5pt}

\paragraph{\modelname Improves General Alignment Ability.}
Following~\citet{wang2023openchat}, we conduct experiments on AlpacaEval, MT-Bench, and Vicuna-Bench to evaluate the general alignment ability.
From Tab.~\ref{tab:alignment_ability}, we can see an increase in almost all three datasets and benchmarks on both the LLaMA-2-13B and Mistral-7B models, showing that \modelname can significantly improve models' response quality and conversation ability. Additionally, from the results of MT-Bench, we can see that \modelname can improve the performance of the second turn of conversations on half the datasets even though it only rewrites the first turn of the instruction data.

% \vspace{-5pt}

\paragraph{\modelname Can Boost Math Reasoning.}
To evaluate the effect of \modelname on math reasoning, we apply \modelname to \texttt{GSM8K} and \texttt{MATH} datasets. As shown in Tab.~\ref{tab:math_reasoning}, \modelname can dramatically boost the math reasoning on both datasets, even up to 9.86\% on \texttt{GSM8K} using LLaMA-2-13B.
Remarkably, \modelname enhances generalization, demonstrated by cross-domain performance boosts. Specifically, training models using the \texttt{MATH} dataset yields notable improvements in the \texttt{GSM8K} test results, and vice versa. For instance, it has been observed that training on the \texttt{MATH} dataset can augment \texttt{GSM8K} performance by 10.69\% based on LLaMA-2-13B.
We explore the possible reasons in \S\ref{sec:math_insights}.
% One possible reason is that \modelname can provide more and clearer intermediate steps and explanations, thereby bolstering the reasoning ability of models.

\begin{figure}[!t]
	\centering
	\subfigure[GPT-4 Evaluation]{
		\begin{minipage}[b]{0.48\textwidth}
			\includegraphics[width=1\textwidth]{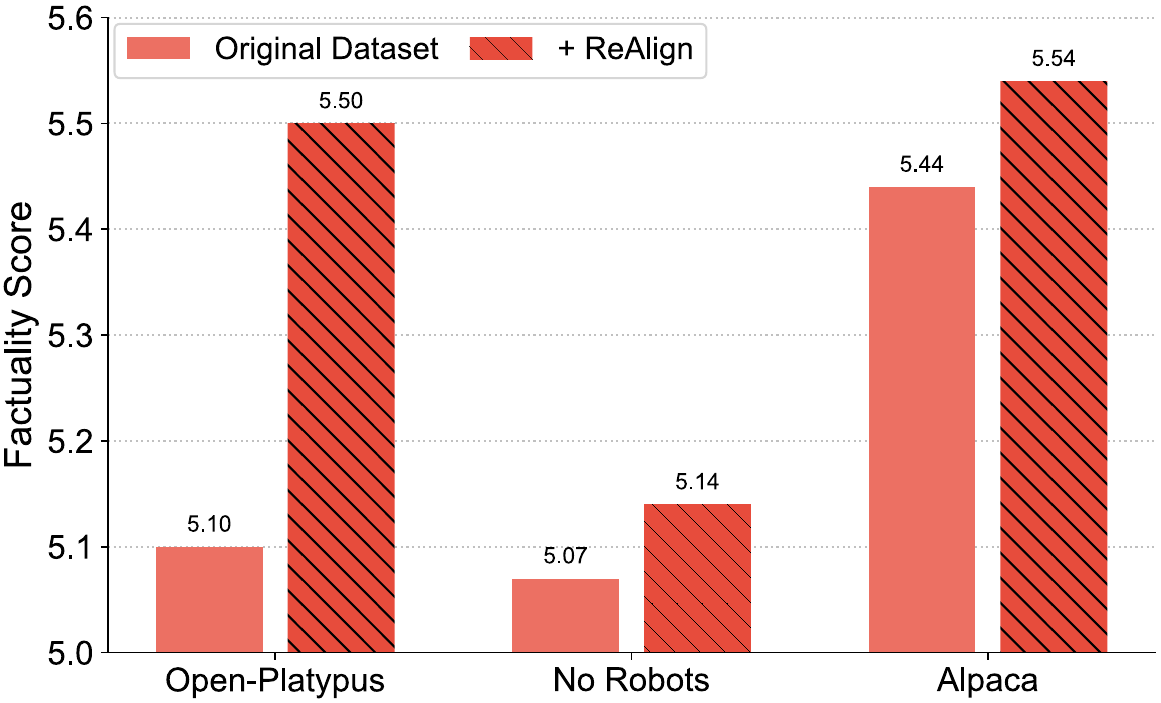} 
		\end{minipage}
		\label{fig:factscore_gpt4}
	}
        % \hspace{20pt}
    	\subfigure[Human Evaluation]{
    		\begin{minipage}[b]{0.48\textwidth}
   		 	\includegraphics[width=1\textwidth]{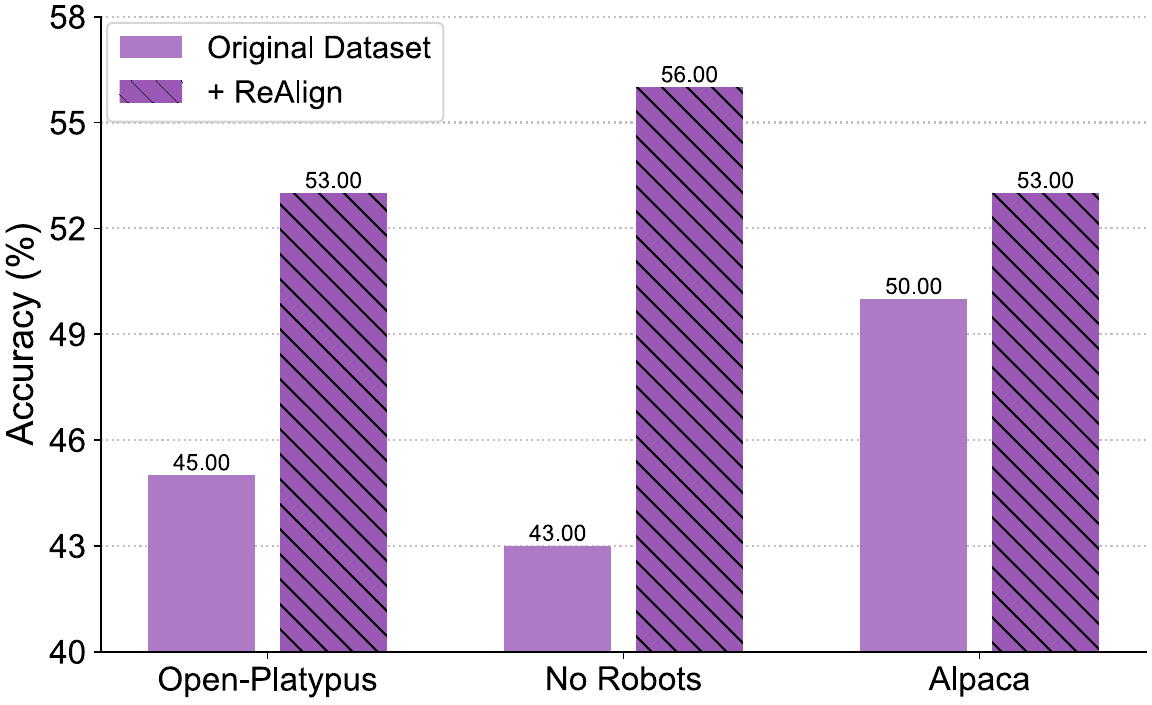}
    		\end{minipage}
		\label{fig:factscore_human}}
     % \vspace{-8pt}
	\caption{The results of the factuality score evaluated by GPT-4 and human.}
 % \vspace{-15pt}
	\label{fig:factscore}
\end{figure}

% \vspace{-5pt}
\paragraph{\modelname Can Enhance Factuality.}
\label{sec:factuality}
To evaluate the factuality, we employ \modelname to \texttt{Open-Platypus}, \texttt{No Robots}, and \texttt{Alpaca} datasets with LLaMA-2-13B, subsequently comparing the response to ground truth in NQ samples.
Fig.~\ref{fig:factscore} shows \modelname elevates the factuality, highlighting its efficacy. This improvement is probably due to the addition of retrieval augmentation.

% \begin{figure}[!t]
% 	\centering
% 	\includegraphics[width=0.6\columnwidth]{Graph/draw_fact_nq.pdf}
%  % \vspace{-8pt}
% 	\caption{The results of the factuality score.}
%  % \vspace{-20pt}
% 	\label{fig:factscore}
% \end{figure}

\begin{figure}[!t]
	\centering
	\subfigure[GPT-4 Judgments]{
		\begin{minipage}[b]{0.48\columnwidth}
			\includegraphics[width=1\columnwidth]{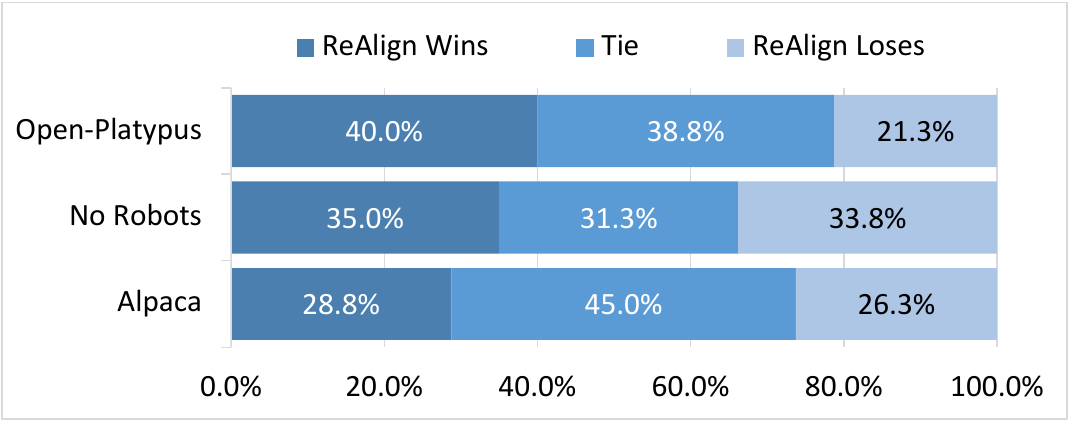} 
		\end{minipage}
		\label{fig:readability_gpt4_results}
	}
    	\subfigure[Human Judgments]{
    		\begin{minipage}[b]{0.48\columnwidth}
		 	\includegraphics[width=1\columnwidth]{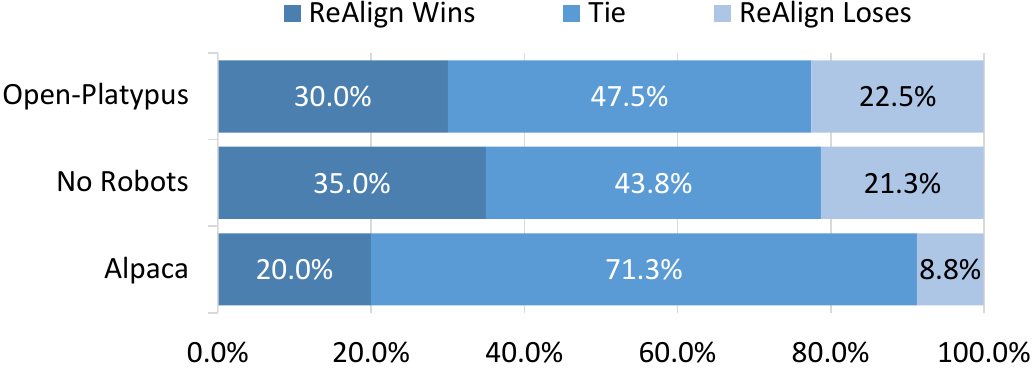}
    		\end{minipage}
		\label{fig:readability_human_results}
    	}
     % \vspace{-12pt}
	\caption{The readability win-rate of the original dataset + \modelname against the original dataset based on LLaMA-2-13B, judged by GPT-4 and human.}
 % \vspace{-12pt}
	\label{fig:readability_results}
\end{figure}

% \vspace{-5pt}

\paragraph{\modelname Can Improve Readability.}

% \begin{table}[t]
\begin{wraptable}{r}{8cm}

  \centering
  % \scriptsize
  \setlength{\tabcolsep}{4pt}
    \begin{tabular}{l|cc}
    \toprule
    Dataset & Response Len. & \modelname \% \\
    \midrule
    \texttt{Open-Platypus} & 224.92 & - \\
    \ \ + \modelname & 206.91 & 28.5\% \\
    \midrule
    \texttt{No Robots} & 211.99 & - \\
    \ \ + \modelname & 211.54 & 15.9\%\\
    \midrule
    \texttt{Alpaca} & 65.51 & - \\
    \ \ + \modelname & 72.38 & 29.9\% \\
    \midrule
    \texttt{GSM8K} & 130.59 & - \\
    \ \ + \modelname & 327.65 & 100\% \\
    \midrule
    \texttt{MATH} & 243.73 & - \\
    \ \ + \modelname & 375.35 & 100\% \\
    \bottomrule
    \end{tabular}%
    % \vspace{-5pt}
  \caption{The datasets analysis includes original datasets and them + \modelname. Response Len. is the average number of tokens of the responses. \modelname \% denotes the percentage of successful reformatting after the adaptive rewriting.} 
  % \vspace{-15pt}
  \label{tab:dataset_analysis}
  \end{wraptable}
% \end{table}

To evaluate the readability of the responses, we use a readability evaluation prompt (refer to Tab.~\ref{tab:readability_evaluation_prompt}) to guide GPT-4 and human to compare the model trained on the original dataset with the model trained with the addition of \modelname. As shown in Fig.~\ref{fig:readability_results}, we see that \modelname can improve the readability of three datasets, especially in the \texttt{Open-Platypus} dataset (i.e., 18.7\% improvements in GPT-4 judgments). It demonstrates that designing different formats for different tasks and reformatting them can effectively improve readability. In addition, human tends to provide more ties for judgments compared to GPT-4. A possible reason is that \modelname can provide better structure, causing GPT-4 to be limited to surface formats ignoring content and deep structure. In contrast, humans can read more carefully not being limited to surface formats.

\begin{figure}[t]
	\centering
	\subfigure[BBH]{
		\begin{minipage}[b]{0.48\columnwidth}
			\includegraphics[width=1\columnwidth]{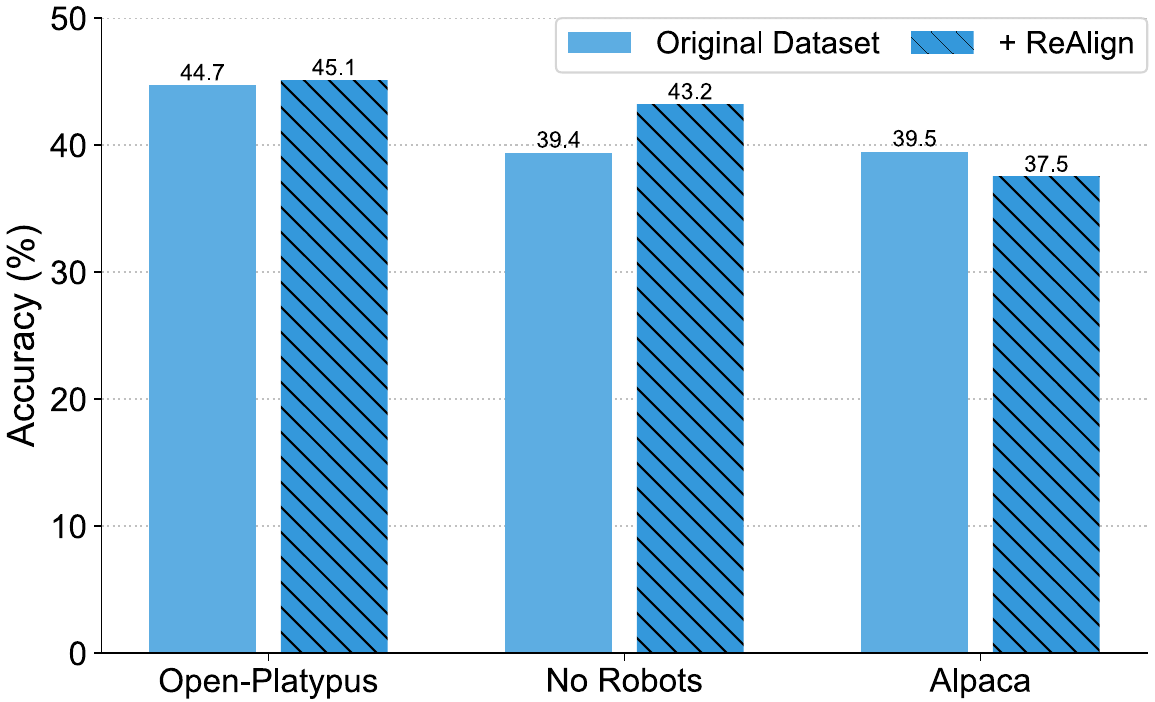} 
		\end{minipage}
		\label{fig:reasoning_ability_bbh}
	}
        % \hspace{20pt}
    	\subfigure[AGIEval]{
    		\begin{minipage}[b]{0.48\columnwidth}
   		 	\includegraphics[width=1\columnwidth]{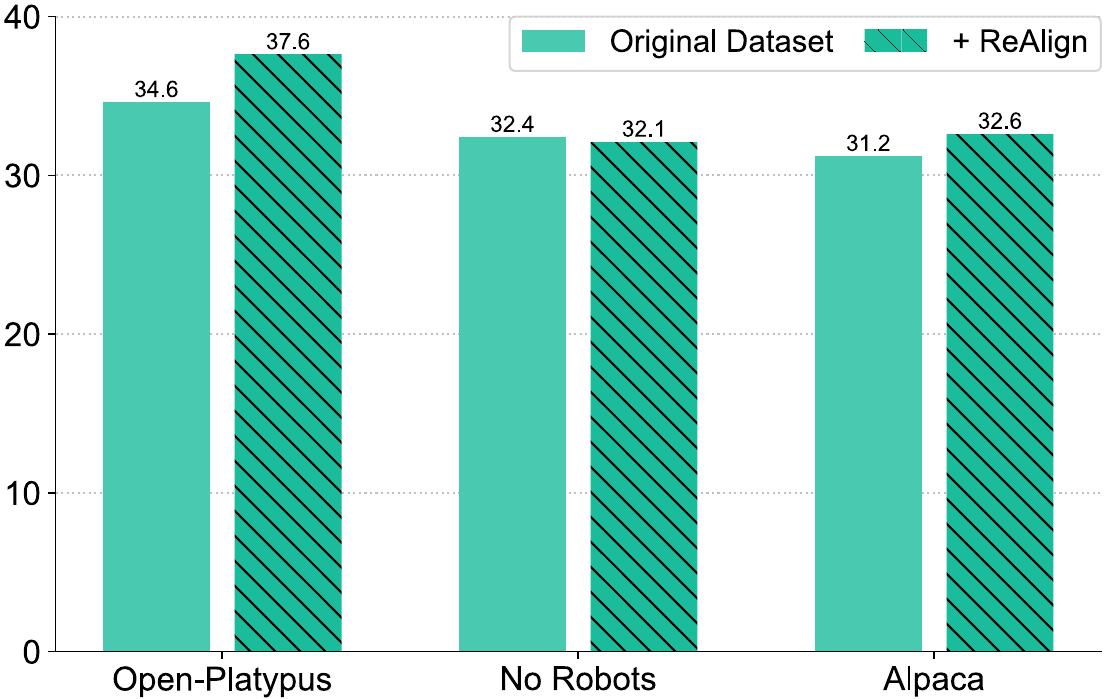}
    		\end{minipage}
		\label{fig:reasoning_ability_agieval}
    	}
     % \vspace{-8pt}
	\caption{The results of the knowledge abilities, including the Big Bench Hard (BBH) (3-shot), and AGIEval (zero-shot). We evaluate the abilities across the \texttt{Open-Platypus}, \texttt{No Robots}, and \texttt{Alpaca} datasets, based on LLaMA-2-13B.}
 % \vspace{-10pt}
	\label{fig:reasoning_ability}
\end{figure}

% \vspace{-3pt}
\subsection{Analysis}
% \vspace{-2pt}
\subsubsection{Datasets Analysis}
% \vspace{-2pt}

First, we compare the change in the length of responses (i.e., the number of tokens) between the original datasets and the addition of \modelname, finding that \texttt{Open-Platypus} becomes shorter and \texttt{No Robots} does not change much, while \texttt{Alpaca}, \texttt{GSM8K}, and \texttt{MATH} become longer (see Tab.~\ref{tab:dataset_analysis}).
Second, we calculate the percentage of responses for which the adaptive rewriting method selects rewrite by edit distance (the results are shown in Tab~\ref{tab:dataset_analysis}). Specifically, we compute the edit distance (including substitution, deletion, and insertion) on a word basis, then divide the edit distance by the length of the longest of the original and rewritten responses to obtain the edit rate, and finally record those with an edit rate greater than 0.2 as rewritten, and the rest as unrewritten. For \texttt{GSM8K} and \texttt{MATH} datasets, all data are ReAligned as adaptive rewriting was not applied to them.

% \vspace{-5pt}
\subsubsection{Why Does \modelname Boost Math?}
\label{math_reason}

% \vspace{-5pt}
A series of experiments and analyses yield several important insights (see the complete version in \S\ref{sec:math_insights}):
\paragraph{(1) A well-organized format is more beneficial than merely providing step-by-step explanations.} As shown in Tab.~\ref{tab:math_explanation_exp}, merely providing a step-by-step explanation is insufficient without a well-organized format.
% \vspace{-5pt}
\paragraph{(2) Length is not all you need.} To analyze the impact of the length of reasoning steps, we hypothesize that the longer the response, the more extensive the reasoning steps involved. As shown in Tab.~\ref{tab:math_explanation_exp} and Tab.~\ref{tab:math_data_length}, length is not the determining factor; rather, a well-organized format can lead to more substantial gains.
% \vspace{-5pt}
\paragraph{(3) Human value is the most important principle in designing formats.} As shown in Tab.~\ref{tab:math_variants_exp}, formats that align with human habits and are easier to understand yield better performance. Therefore, we also advocate that the development of large language models should move closer to user values.

% \vspace{-5pt}
\subsubsection{Alignment Tax}

% \begin{table}[t]
\begin{wraptable}{r}{8cm}

  \centering
  % \scriptsize
  \setlength{\tabcolsep}{4pt}
    \begin{tabular}{l|ccc}
    \toprule
    Dataset & General Align. & Know. Ab. & FS \\
    \midrule
    \texttt{Open-Platypus} & 6.18 & 39.65 & 5.1 \\
    \ + \modelname & \textbf{6.24} & \textbf{41.35} & 5.5 \\ \midrule
    \ \ W/o RA & 6.18 & 40.6 & 5.3 \\
    \ \ W/o Adaption & 6.17 & 39.8 & \textbf{5.6} \\
    \bottomrule
    \end{tabular}%
    % \vspace{-5pt}
  \caption{Ablation study results show that removing retrieval augmentation is indicated by "W/o RA" and removing adaptive rewriting by "W/o Adaption" in \modelname. ``General Align.'' and ``Know. Ab.'' denotes general alignment ability and Knowledge Ability, which are the average results. FS denotes Factuality Score. \textbf{Bold} denotes the best.}
  % \vspace{-15pt}
  \label{tab:ablation_study}
  \end{wraptable}
% \end{table}

When the model is fine-tuned on the \modelname dataset, a question worth exploring is whether there is a drop in knowledge ability even as alignment ability improves.
To evaluate the knowledge ability, we follow~\cite{mitra2023orca} to employ the following benchmarks: \textbf{Big Bench Hard (BBH)}~\cite{suzgun2022challenging} and \textbf{AGIEval}~\cite{zhong2023agieval}, which is multiple choices knowledge-intensive QA task.
As shown in Fig.~\ref{fig:reasoning_ability}, we can see that \modelname has little effect on the knowledge-based tasks, indicating that our approach does not impair the knowledge in the original dataset. 
It is worth noting that in some cases \modelname will also provide a significant boost to knowledge, such as \texttt{Open-Platypus} on AGIEval. Possible reasons are that a well-defined format can facilitate the accuracy of the knowledge-based tasks~\cite{wei2022cot} and that retrieving external information can augment knowledge.

\begin{figure}[t]
	\centering
	\includegraphics[width=0.7\columnwidth]{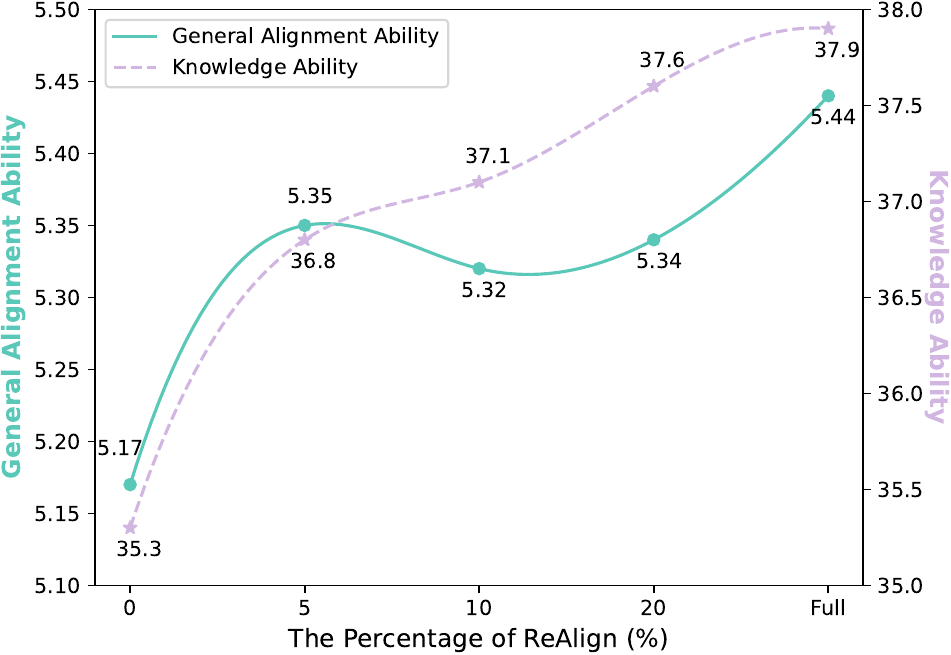}
 % \vspace{-5pt}
	\caption{The scaling trends in \modelname data percentage, including general alignment ability and knowledge ability. We conduct the experiment in the \texttt{Alpaca} dataset based on LLaMA-2-13B.}
 % \vspace{-15pt}
	\label{fig:scaling_law}
\end{figure}
% \vspace{-5pt}
\subsubsection{Ablation Studies}
% \vspace{-5pt}
We rewrite two variants of the \texttt{Open-Platypus} dataset and train them based on LLaMA-2-13B for ablation studies:

\begin{enumerate}[label={(\arabic*)}, wide, labelwidth=!, labelindent=0pt] 
% \vspace{-6pt}
    \item \textbf{W/o Retrieval Augmentation}: We remove the retrieval augmentation from \modelname and rewrite all tasks without evidences. As shown in Tab.~\ref{tab:ablation_study}, the general alignment ability, knowledge ability, and factuality score (FS) are reduced, indicating the effectiveness of retrieval augmentation. Interestingly, the FS metrics are higher without RA than in the original dataset, suggesting that \modelname also has the potential to improve the factuality.
    % \vspace{-6pt}
    \item \textbf{W/o Adaption}: We remove the adaptive rewriting from \modelname and use force rewriting. Tab.~\ref{tab:ablation_study} shows the general alignment and knowledge ability decrease. This may be because forced rewriting, while making the responses more structured, ignores the question's requirements, weakening the instruction-following ability. In addition, FS has increased, probably because forced rewriting leads to a larger amount of \modelname data, introducing more retrieved knowledge and boosting factuality.
    % \vspace{-5pt}
\end{enumerate}

% \vspace{-10pt}
\subsubsection{The Scaling Law of \modelname}
% \vspace{-3pt}
We experiment to explore the impact of the number of \modelname. Specifically, we randomly sample a $k\%$ ($k=0, 5, 10, 20, \text{Full}$, with $\text{Full}$ being $29.9\%$) of \modelname \texttt{Alpaca} data, and fill in the remainder with original responses. The original Alpaca dataset corresponds to $0\%$.
Interestingly, we find that only 5\% of the \modelname data yields a 67\% boost in general alignment ability compared to the entire \modelname data (see Fig.~\ref{fig:scaling_law}). 
This suggests that only a small amount of data is required to learn style and format, to expose the knowledge and capabilities that were already acquired during pretraining~\cite{zhou2023lima}. Additionally, the knowledge capability continues to improve as the amount of \modelname data improves.

% \vspace{-5pt}
\subsubsection{Case Study}
% \vspace{-3pt}
We show a case from the MT-Bench test set in Tab.~\ref{tab:case-study}. This example shows that the response given by the \modelname model has a better format.

\section{Conclusion}
In this work, we propose \modelname, a simple and effective method for alignment, which automatically improves the quality of the existing instruction datasets while minimizing labor costs and hallucinations. We create five new high-quality datasets from \texttt{Open-Platypus}~\cite{lee2023platypus}, \texttt{No Robots}~\cite{no_robots}, \texttt{Alpaca}~\cite{alpaca}, \texttt{GSM8K}~\cite{cobbe2021gsm8k}, and \texttt{MATH}~\cite{hendrycksmath2021} and high-quality manual-written natural language formats. Experiments demonstrate that \modelname significantly boosts general alignment ability, math reasoning, factuality, and readability without impairing knowledge ability. Last, we release the code, defined criteria, and data to facilitate future research.

\section*{Limitations}
First, our approach relies on the ability to reformatting models, which is currently less effective in open-source models (e.g., LLaMA2~\cite{touvron2023llama2}) but more costly in closed-source models (e.g., GPT-4~\cite{openai2023gpt4}).
Second, the task categories we define cannot cover all tasks in reality, as real questions may be more complex and involve multiple tasks. Therefore, it is necessary to define more tasks and formats for a wide range of diverse and regional scenarios.
Third, applying \modelname only to single-turn conversations has the potential to hurt the alignment ability of the second-turn conversations, hence extending \modelname to multi-turn conversation would also be valuable.
Last, we explore the reasons behind \modelname's success superficially, and thus will further explore the science and mechanistic interpretability behind it in the future.

\section*{Ethics Statement}
We take ethical considerations very seriously. In this paper, both the datasets and models are publicly available and have been widely adopted by researchers. We ensure that the findings and conclusions of this paper are reported accurately and objectively.

\section*{Acknowledgements}
We thank the GAIR members for reviewing our paper and giving valuable feedback. We appreciate the authors in~\citet{wang2023openchat} for providing the training codebase and the helpfulness.

\bibliography{main}
\bibliographystyle{acl_natbib}

\appendix
\section{Implementation Details}
For retrieval augmentation, we select the top-$5$ evidence from the retrieval results. For reformatting, We guide gpt-3.5-turbo-1106 to reformat the responses. We set the temperature as $0.3$, the top-p as $0.1$, and the target length as $2048$. Additionally, we generate two rewrite results at a time and choose the longest one, hence $n$ is set to $2$.
For training, we fine-tune the models based on the LLaMA-2-13B~\cite{touvron2023llama2} and {Mistral-7B}~\cite{jiang2023mistral} for 5 epochs on the \texttt{Open-Platypus} dataset, 20 epochs on the \texttt{No Robots} and \texttt{Alpaca} datasets, and 3 epochs on the \texttt{GSM8K} and \texttt{MATH} datasets, using the AdamW optimizer with a sequence length of 4,096 tokens. The batch size is 64 for the \texttt{Open-Platypus}, \texttt{No Robots}, and \texttt{Alpaca} datasets, and 128 for the \texttt{GSM8K} and \texttt{MATH} datasets. The AdamW optimizer's hyperparameters are set as follows: $\beta_1=0.9$, $\beta_2=0.95$, $\epsilon=10^{-5}$, and weight decay of $0.1$. We employ a cosine learning rate schedule with a maximum learning rate of $5.7 \times 10^{-5}$ for the \texttt{Open-Platypus} dataset, $6.25 \times 10^{-5}$ for the \texttt{No Robots} dataset, $6.55 \times 10^{-5}$ for the \texttt{Alpaca} dataset, and $1 \times 10^{-5}$ for the \texttt{GSM8K} and \texttt{MATH} datasets, which decays to 10\% of the maximum value. Following~\citet{wang2023openchat} and~\citet{DBLP:journals/jmlr/GranziolZ022}, the learning rate is scaled proportionally to the square root of the batch size. All models are trained on 8 NVIDIA A100 80G GPUs.

\section{Task Description}
\label{sec:tasks_description}
The task descriptions mentioned in \S\ref{sec:mothed_tasks} and whether they are retrieved and rewritten are exhibited in Tab.~\ref{tab:tasks_description}.

\section{Training Details of Task Classifier}
\label{sec:task_classifier}
In this section, we describe the training details of the task classifier mentioned in \S\ref{sec:mothed_tasks}.

% \begin{table}[t]
\begin{wraptable}{r}{9cm}
\small
  \centering
  % \scriptsize
  \setlength{\tabcolsep}{1pt}
    \fontsize{9}{12}\selectfont
    \begin{tabular}{c|lccc}
    \toprule
    Test & \multicolumn{1}{c}{Training} & \multicolumn{2}{c}{Model} & \multirow{2}{*}{Overall} \\
    \cmidrule(lr){3-4} Dataset & \multicolumn{1}{c}{Dataset} & LLaMA-2-13B & Mistral-7B & \\\midrule

    \multirow{3}{*}{GSM8K} & GSM8K & 46.77 & 61.25 & 54.01 \\
                           & \ + Explanation & 48.60 & 53.37 & 50.99 \\
                           & \ + \modelname & \textbf{56.63} & \textbf{68.16} & \textbf{62.40}\\
    \midrule
    \multirow{3}{*}{MATH} & MATH & 6.14 & 13.18 & 9.66 \\
                           & \  + Explanation & \textbf{7.30} & 13.94 & 10.62 \\
                           & \ + \modelname & 7.14 & \textbf{15.30} & \textbf{11.22} \\
    \bottomrule
    \end{tabular}%
    % \vspace{-5pt}
  \caption{The results of math reasoning on \texttt{GSM8K}, \texttt{MATH} and them + explanation or \modelname. We report the accuracy by exact matching. \textbf{Bold} indicates the best result.}.
  % \vspace{-25pt}
  \label{tab:math_explanation_exp}
  \end{wraptable}
% \end{table}

In real scenarios, user questions can be quite long and complex (with more than 1k words), while traditional BERT-like models only have a context length of 512 tokens, urging us to train a large language model for classification.
Following~\citet{li2023generative}, we convert the classification task into a generation task, which directly generates the task name given a question with the prompt as shown in Tab.~\ref{tab:classification_prompt}. Specifically, we manually label about 33 questions for each kind of task from \texttt{LIMA}~\cite{zhou2023lima}, \texttt{ShareGPT}~\cite{zheng2023judging}, and \texttt{Alpaca}~\cite{alpaca} datasets. For tasks with less training data, we use ChatGPT to generate a portion of the questions.
Then, we divide questions in a 9:1 train/test split (Tab.~\ref{tab:task-cls-data-distribution}). We train the task classifier from LLaMA-2-13B~\cite{touvron2023llama2}, and set the max sequence length as 2,048, epochs as 20, and batch size as 16. We set the initial learning rate to 2e-5 and consine decaying to 0 by the end of training with warmup steps as 10. The optimizer is AdamW with $\beta_1=0.9, \beta_2=0.95$. The loss is only calculated on the output end as well. The accuracy and F1 of the final task classifier on the test set are 78.32\% and 81.59\%, respectively.

\section{The Description of Auto-J}
\label{sec:description_autoj}
Auto-J~\cite{li2023generative} is an open-source generative judge designed to evaluate LLMs based on their alignment with human preferences, which is the best critique model besides GPT-4~\cite{sun2024metacritique}. Auto-J stands out due to its generality, being trained on real-world user queries and responses from various LLMs across 58 scenarios. It offers flexibility by enabling both pairwise comparison and single-response evaluation through prompt adjustments. Additionally, Auto-J enhances reliability and encourages human participation in the evaluation process by offering detailed natural language critiques, improving interpretability.

\section{Why Does \modelname Boost Math? (Complete Version)}
\label{sec:math_insights}
By carefully observing the cases, we speculate that the reason for the improvements in math reasoning may stem from the easier-to-understand format, more detailed explanations~\cite{mukherjee2023orca}, or length~\cite{jin2024impact}. To further explore the reasons, we merely incorporate detailed step-by-step explanations without including a complete format. These explanations are generated using gpt-3.5-turbo-1106, with the prompts used detailed in Tab.~\ref{tab:generation_prompts_explanation_gsm8k} and Tab.~\ref{tab:generation_prompts_explanation_math}. The results are shown in Tab.~\ref{tab:math_explanation_exp}.

Moreover, to explore the impact of various formats, we experiment with two other formats based on \modelname: (1) The first requires separating natural language and calculation, meaning that the natural language does not include the calculation process and mathematical computations are expressed separately; (2) The second variant requires the use of special markers `$<<>>$' in equations based on the format of \modelname, for example, $<<1+2=3>>$. We merely create these variants on GSM8K due to the cost of ChatGPT API. The results are shown in Tab.~\ref{tab:math_variants_exp}

% (1) We merely incorporate detailed step-by-step explanations without including a complete format. These explanations are generated using gpt-3.5-turbo-1106, with the prompts used detailed in Tab.~\ref{tab:generation_prompts_explanation_gsm8k} and Tab.~\ref{tab:generation_prompts_explanation_math}; (2) We experiment with another format, which requires separating natural language and numbers, meaning that the natural language does not include the calculation process and mathematical computations are expressed separately; (3) We experiment with the second variant with different format, which requires the use of special markers `$<<>>$' in equations based the format of \modelname, for example, $<<1+2=3>>$.

% \begin{table}[t]
\begin{wraptable}{r}{9.2cm}
\small
  \centering
  % \scriptsize
  \setlength{\tabcolsep}{1pt}
    \fontsize{9}{12}\selectfont
    \begin{tabular}{l|ccc}
    \toprule
    \multirow{2}{*}{Dataset} & \multicolumn{2}{c}{Model} & \multirow{2}{*}{Overall} \\
    \cmidrule(lr){2-3} & LLaMA-2-13B & Mistral-7B & \\\midrule
    GSM8K & 46.77 & 61.25 & 54.01 \\
    \ + \modelname (Separate.) & 55.57 & 62.09 & 58.83 \\
    \ + \modelname (`$<<>>$') & \textbf{57.01} & 63.61 & 60.31 \\
    \ + \modelname & 56.63 & \textbf{68.16} & \textbf{62.40} \\
    \bottomrule
    \end{tabular}%
    % \vspace{-5pt}
  \caption{The results of math reasoning on \texttt{GSM8K} and them + different formats of \modelname. We report the accuracy by exact matching. ``Separate.'' denotes the first variant that separates natural language and calculation. ``$<<>>$'' denotes the second variant that requires the use of special markers `$<<>>$' in equations based on the format of \modelname. \textbf{Bold} indicates the best result.}.
  % \vspace{-25pt}
  \label{tab:math_variants_exp}
  \end{wraptable}
% \end{table}

From the results of the above experiments (see Tab.~\ref{tab:math_explanation_exp} and Tab.~\ref{tab:math_variants_exp}), we can derive insights below:
\paragraph{Insights 1: A well-organized format is more beneficial than merely providing step-by-step explanations.} As shown in Tab.~\ref{tab:math_explanation_exp}, we can find that merely providing a detailed explanation is insufficient without a well-structured format and may even result in performance inferior to the original dataset (i.e., The results on GSM8K based on Mistral-7B). For the more complex MATH dataset, detailed explanations still play a significant role. However, A well-organized structure may further enhance their effectiveness.

\paragraph{Insights 2: Length is not all you need.} \citet{jin2024impact} suggests that longer reasoning steps can enhance math reasoning capabilities. To further analyze the impact of the length of reasoning steps, we hypothesize that the longer the response, the more extensive the reasoning steps involved. Specifically, we calculate the average length that is shown in Tab.~\ref{tab:math_data_length}. We can see that the average length of ``GSM8K + Explanation'' exceeds that of ``GSM8K'' by more than double, and is even longer than ``GSM8K + ReAlign''. However, its performance is significantly inferior to ``GSM8K + ReAlign''. Additionally, the average length of "MATH + Explanation" is shorter than "MATH + ReAlign", yet it demonstrates superior performance on LLaMA-2-13B. These findings suggest that length is not the determining factor; rather, a well-organized format can lead to more substantial gains.

% \begin{table}[t]
\begin{wraptable}{r}{5cm}
\small
  \centering
  % \scriptsize
  \setlength{\tabcolsep}{4pt}
    \begin{tabular}{lc}
    \toprule
    Dataset & Response Len. \\
    \midrule
    \texttt{GSM8K} & 130.59  \\
    \ + Explanation & 341.61 \\
    \ + \modelname & 327.65 \\
    \midrule
    
    \texttt{MATH} & 243.73 \\
    \ + Explanation & 293.57 \\
    \  + \modelname & 375.35 \\
    \bottomrule
    \end{tabular}%
  \caption{The response length of original datasets and them + Explanation and \modelname. Response Len. is the average number of tokens of the responses.} 
  \label{tab:math_data_length}
  \end{wraptable}
% \end{table}

\paragraph{Insights 3: Human value is the most important principle in designing formats.} As shown in Tab.~\ref{tab:math_variants_exp}, we can see that, firstly, expressing natural language and mathematical calculations together (\modelname) performs better. This approach is more in line with human habits and preferences, making it easier for users to understand. Secondly, adding special markers ``$<<>>$'' to equations can diminish performance, potentially because this format does not align with the customary practices of human writing and usage, making it less intuitive to understand.
Therefore, the most important principle in designing formats is human value. A good format is not only about superior performance but also about user preference (i.e., formats that are easier for users to understand). Following this principle in designing criteria, we observed a significant improvement in our math reasoning abilities, indicating that formats easier for users to understand can also lead to better performance. Therefore, we also advocate that the development of large language models should move closer to user values.

\section{Prompts}
The reformatting prompts for non-knowledge-intensive tasks and knowledge-intensive tasks are shown in Tab.~\ref{tab:rewriting_prompts_non_retrieval} and Tab.~\ref{tab:rewriting_prompts_retrieval}. The complete version of Fig.~\ref{fig:example} and Tab.~\ref{tab:format_example} are shown in Tab.~\ref{tab:complete_example} and Tab.~\ref{tab:complete_format}, respectively. The example of a knowledge-intensive task is shown in Tab.~\ref{tab:kilt_example}. The evaluation prompts for readability and factuality are shown in Tab.~\ref{tab:readability_evaluation_prompt} and Tab.~\ref{tab:factuality_evaluation_prompt}, respectively. The prompts for generating explanations for GSM8K and MATH datasets are shown in Tab.~\ref{tab:generation_prompts_explanation_gsm8k} and Tab.~\ref{tab:generation_prompts_explanation_math}.

\begin{table*}[t]
\small
  \centering
  \scriptsize
    \begin{tabular}{p{62em}}
    \toprule
    \textbf{Query}: Benjamin went to a bookstore and purchased a variety of books. He bought $5$ copies of a sci-fi novel, each priced at \$20, 3 copies of a history book priced at \$30 each, and 2 copies of a philosophy book for \$45 each.
    What was the total cost of his purchases? \\
    \midrule
    \textbf{Response from Original Model}: To find the total cost of Benjamin's purchases, I need to add up the prices of all the books he bought. I can use the information given in the problem to set up an equation:
    
    \redback{Total cost = Price of sci-fi novel * Number of copies + Price of history book * Number of copies + Price of philosophy book * Number of copies}

    \redback{Plugging in the values from the problem, I get:}

    \redback{Total cost = \$20 * 5 + \$30 * 3 + \$45 * 2}

    \redback{Simplifying, I get:}

    \redback{Total cost = \$100 + \$90 + \$90}

\redback{Adding, I get:}

\redback{Total cost = \$280}

So, the total cost of Benjamin's purchases was \$280.  \\
    \midrule
    \textbf{Response from \modelname Model}:
To find the total cost of Benjamin's purchases, we need to add up the prices of the books he bought.
\\\\
\greenback{Step-by-step solution:}

\greenback{1. Calculate the cost of the sci-fi novels: 5 copies at \$20 each = \$100}

\greenback{2. Calculate the cost of the history books: 3 copies at \$30 each = \$90}

\greenback{3. Calculate the cost of the philosophy books: 2 copies at \$45 each = \$90}
\\\\
\greenback{Detailed explanation:}

\greenback{The total cost of Benjamin's purchases is the sum of the costs of the sci-fi novels, the history books, and the philosophy books: \$100 + \$90 + \$90 = \$280.}
\\\\
\greenback{Correct result and conclusion:}

The total cost of Benjamin's purchases was \$280.
     \\
    \bottomrule
    \end{tabular}%
  \caption{An example of the response from a model trained on the original \texttt{Open-Platypus} dataset (denoted as Original Model) and the response from a model trained on the \modelname \texttt{Open-Platypus} dataset (denoted as \modelname Model). We highlight the content in the better format given by the \modelname Model in \greenback{green} compared with the content given by the Original Model in \redback{red}.}
  \label{tab:case-study}
\end{table*}

\begin{table*}[!htbp]
  \centering
  \tiny
    \begin{tabular}{c|l|c|c}
    \toprule
    \textbf{Task name} & \multicolumn{1}{c}{\textbf{Description}} & \textbf{Retrieval} & \textbf{Rewriting} \\
    \midrule
    \multicolumn{4}{c}{\textit{Generation}} \\
    \midrule
    question generation & Write some questions based on the given description. & \XSolidBrush & \Checkmark \\
    story generation & Write a story based on the given description. & \XSolidBrush & \XSolidBrush \\
    poem generation & Write a poem based on the given description. & \XSolidBrush & \XSolidBrush  \\
    email generation & Write an email based on the given description. & \XSolidBrush & \Checkmark  \\
    data generation & Generate data based on the given description. & \XSolidBrush &  \Checkmark \\
    text-to-text translation & Translate the given text into another language. & \XSolidBrush & \XSolidBrush \\
    \midrule
    \multicolumn{4}{c}{\textit{Brainstorming}} \\
    \midrule
    advice giving & Respond well to users when they seek advice. & \XSolidBrush & \XSolidBrush \\
    recommendations & Give recommendations to users. & \Checkmark & \Checkmark \\
    how-to generation & Give relevant and complete answer when users ask `how to do` something. & \Checkmark & \Checkmark \\
    planning & Write a plan for an event or activity. & \XSolidBrush & \Checkmark \\
    \midrule
    \multicolumn{4}{c}{\textit{Code}} \\
    \midrule
    code correction & Correct the potential errors in a piece of code. & \XSolidBrush & \Checkmark \\
    code simplification & Rewrite a piece of code to make it more concise and easy to understand. & \XSolidBrush & \XSolidBrush  \\
    explain code & Write an explanation for a piece of code. & \XSolidBrush & \Checkmark \\
    text-to-code translation & Write a piece of code based on the given description. & \XSolidBrush & \Checkmark \\
    code-to-code translation & Convert the given code into another programming language. & \XSolidBrush & \Checkmark \\
    language learning questions & Write an answer for the given question about programming language learning. & \XSolidBrush & \Checkmark \\
    code language classification & Classify the programming language for the given code. & \XSolidBrush & \Checkmark \\
    code-to-text-translation & Write a document for the given code. & \XSolidBrush & \Checkmark \\
    \midrule
    \multicolumn{4}{c}{\textit{Rewriting}} \\
    \midrule
    instructional rewriting & Rewrite a given text with a specific instruction. & \XSolidBrush & \Checkmark \\
    language polishing & Polish a piece of text to make it more fluent, natural, and readable. & \XSolidBrush & \Checkmark \\
    paraphrasing & Paraphrase a given text. & \XSolidBrush & \XSolidBrush \\
    text correction & Correct the potential errors in a piece of text. & \XSolidBrush & \Checkmark \\
    \midrule
    \multicolumn{4}{c}{\textit{Extraction}} \\
    \midrule
    information extraction & Extract one or multiple user-specified categories of information from a piece of text attached in the user's query. & \XSolidBrush & \Checkmark \\
    keywords extraction & Extract the keywords from a piece of text. & \XSolidBrush & \Checkmark \\
    table extraction & Generate a table include the key information from a piece of text attached in the user's query. & \XSolidBrush & \XSolidBrush \\
    \midrule
    \multicolumn{4}{c}{\textit{Summarization}} \\
    \midrule
    title generation & Generate a title for the given text or based on a description of the work. & \XSolidBrush & \XSolidBrush \\
    text summarization & Write a summary for a piece of text. & \XSolidBrush & \XSolidBrush \\
    note summarization & Write a note to summarize a piece of text. & \XSolidBrush & \XSolidBrush \\
    \midrule
    \multicolumn{4}{c}{\textit{Conversation}} \\
    \midrule
    open qa & The user’s query is an open domain question with no attached passage or article. & \Checkmark & \Checkmark \\
    closed qa & Answer the questions that can be directly answered by the attached passage. & \XSolidBrush  & \Checkmark \\
    fact verification & Verify if the given fact is true or false.& \Checkmark & \Checkmark \\
    value judgment & Provide a value judgment on a given topic or statement. & \XSolidBrush & \Checkmark \\
    roleplay & Pretend to be a specific person, character, profession or identity, and complete the required task on this basis. & \XSolidBrush & \XSolidBrush \\
    explain answer & Explain something the user wants to know. & \Checkmark & \Checkmark \\
    \midrule
    \multicolumn{4}{c}{\textit{Specialized Educational Dialog}} \\
    \midrule
    natural language tutor & Write an answer for the given question about natural language learning. & \XSolidBrush & \Checkmark \\
    exam problem tutor & Solve an exam question (like fill-in-the-blank, multiple choice, problem solving, etc) with no math involved. & \XSolidBrush & \Checkmark \\
    ai tutor & Write an answer for the given question about machine learning, artificial intelligence or language model. & \XSolidBrush & \Checkmark \\
    math puzzles & Write an answer with the step-by-step reasoning process for a math question. & \XSolidBrush & \Checkmark \\
    fill in the blank & Complete the missing parts with the most appropriate words to make the text coherent and meaningful. & \XSolidBrush & \Checkmark \\

    \midrule
    \multicolumn{4}{c}{\textit{Classification}} \\
    \midrule
    general classification & Classify one or multiple objects given by the user into the specified categories. & \XSolidBrush & \Checkmark \\
    ordering & Sort some things, according to some criteria. & \XSolidBrush & \Checkmark \\
    sentiment analysis & Identify and categorize the subjective opinions, attitudes, and feelings of the writer towards a particular subject. & \XSolidBrush & \Checkmark \\
    language classification & Classify the language for the given text. & \XSolidBrush & \Checkmark \\
    topic classification & Extract the high-level topics or themes from a given text, i.e., what kind of topics are discussed in the text. & \XSolidBrush & \Checkmark \\

    \midrule
    \multicolumn{4}{c}{\textit{Others}} \\
    \midrule
    rejecting & Reject to respond when the query is beyond capacity or it violates general ethical and legal rules. & \XSolidBrush & \Checkmark \\
    others & You must choose this if none of the other scenarios match the user's query well. & \XSolidBrush & \Checkmark \\
    
    \bottomrule
    \end{tabular}%
  \caption{Detailed description for each task.}
  \label{tab:tasks_description}%
\end{table*}

\begin{table*}
  \centering
  \scriptsize
  \setlength{\tabcolsep}{3pt}
    \begin{tabular}{l|rr|l|rr|l|rr}
    \toprule
    task & \multicolumn{1}{l}{train} & \multicolumn{1}{l|}{test} & task & \multicolumn{1}{l}{train} & \multicolumn{1}{l|}{test} & task & \multicolumn{1}{l}{train} & \multicolumn{1}{l}{test} \\
    \midrule
    question\_generation & 30 & 2 & code\_language\_classification & 30 & 2 & roleplay & 30 & 3 \\
    story\_generation & 30 & 4 & code\_to\_text\_translation & 30 & 3 & explain\_answer & 30 & 4 \\
    poem\_generation & 30 & 3 & instructional\_rewriting & 30 & 4 & natural\_language\_learning\_tutor & 30 & 2\\
    email\_generation & 30 & 3 & language\_polishing & 30 & 2 &exam\_problem\_solving\_tutor & 31 & 2 \\
    data\_generation & 30 & 3 & paraphrasing & 30 & 2 & ml\_ai\_language\_model\_tutor & 30 & 3 \\
    text\_to\_text\_translation & 30 & 3 & text\_correction & 30 & 2 & math\_puzzles & 30 & 6 \\
    advice\_giving & 30 & 4 & information\_extraction & 30 & 3 & fill\_in\_the\_blank & 30 & 3 \\
    recommendations & 30 & 2 & keywords\_extraction & 30 & 2 & general\_classification & 30 & 4 \\
    how\_to\_generation & 30 & 3 & table\_extraction  & 30 & 3 & ordering & 30 & 3\\
    planning & 30 & 2 & title\_generation & 30 & 2  & sentiment\_analysis & 30 & 3\\
    code\_correction & 30 & 5 & text\_summarization & 30 & 5 & language\_classification & 30 & 3 \\
    code\_simplification & 30 & 2 & note\_summarization & 30 & 2  & topic\_classification &  30 & 2\\
    explain\_code & 30 & 2 & open\_qa & 30 & 6 & rejecting & 30 & 3 \\
    text\_to\_code\_translation & 30 & 4 & closed\_qa & 30 & 2 & others & 43 & 8 \\
    code\_to\_code\_translation & 30 & 3 & fact\_verification & 30 & 2 & overall & 1395 & 143 \\
    language\_learning\_questions & 31 & 5 & value\_judgement & 30 & 2   \\
    \bottomrule
    \end{tabular}%
  \caption{The task distribution in the training and test set for task classifier.}
  \label{tab:task-cls-data-distribution}%
\end{table*}%

\begin{table*}[t]
    \scriptsize
    \centering
\begin{tabular}{@{}p{1\columnwidth}@{}}
\toprule
\textbf{Classification Prompt}\\
You will receive a user's query. Additionally, you are given some pre-defined tasks below:

\\\\

[Existing tasks start]

question\_generation

story\_generation

poem\_generation

email\_generation

data\_generation

advice\_giving

recommendations

how\_to\_generation

planning

instructional\_rewriting

language\_polishing

paraphrasing

text\_correction

code\_correction

code\_simplification

information\_extraction

keywords\_extraction

table\_extraction

title\_generation

text\_summarization

note\_summarization

explain\_code

explain\_answer

text\_to\_text\_translation

text\_to\_code\_translation

code\_to\_code\_translation

code\_to\_text\_translation

open\_qa

closed\_qa

fill\_in\_the\_blank

fact\_verification

math\_puzzles

language\_learning\_questions

natural\_language\_learning\_tutor

exam\_problem\_solving\_tutor

ml\_ai\_language\_model\_tutor

general\_classification

ordering

sentiment\_analysis

code\_language\_classification

language\_classification

topic\_classification

value\_judgement

rejecting

roleplay

default

[Existing tasks end]

\\\\

Your objective is to choose the most appropriate task that can reflect the high-level intention of this query. You should first clearly give out your choice. Your choice should exactly match one of the task names provided above, without any modification. Do not include the task description in your choice.

\\\\

Your output should be just the task name.

\\\

User's query is below:

[User's query start]

\{input\}

[User's query end]

\\\\

Task name:
\\
\bottomrule
\end{tabular}
    \caption{The classification prompt for the task classifier in the training and inference phase.}
    \label{tab:classification_prompt}
\end{table*}

\begin{table*}[t]
    \scriptsize
    \centering
\begin{tabular}{@{}p{1\columnwidth}@{}}
\toprule
\textbf{System Prompt}\\
Please act as a rewriter to modify the format of the AI assistant's response to the user's question presented below.\\\\

Please follow the instructions below:
\\\\

1. Please first determine whether the given format meets the requirements of the user's question, if it does not, then copy the AI assistant's response, if it does, then modify the response's format following the provided format.

2. Your task is limited to altering the format while keeping the original meaning and information intact.

3. Please make sure that the revised response can answer the user's question correctly.

4. Please make sure that the revised response is fluent and has no additional subheadings.
\\\\
Please first write "Reasoning: $<$reason$>$" to provide a brief reasoning you used to modify, and then write "Revised response: $<$response$>$" to output your final revised response without any additional information, ensuring its fluency.
\\\\
Do not output any additional subheadings.\\
\midrule
\textbf{User Prompt}\\
Below is a user's question, the AI assistant's response, and the provided format. 
\\\\

[Question start]

\{question\}

[Question end]

\\\\

[Response start]

\{response\}

[Response end]

\\\\

[Format start]

\{format\}

[Format end] \\
\bottomrule
\end{tabular}
    \caption{Reformatting prompts for non-knowledge-intensive tasks.}
    \label{tab:rewriting_prompts_non_retrieval}
\end{table*}

\begin{table*}[t]
    \scriptsize
    \centering
\begin{tabular}{@{}p{1\columnwidth}@{}}
\toprule
\textbf{System Prompt}\\
Please act as a rewriter to modify the format of the AI assistant's response to the user's question presented below.\\\\

Please follow the instructions below:
\\\\

1. Please first determine whether the given format meets the requirements of the user's question, if it does not, then copy the AI assistant's response, if it does, then modify the response's format following the provided format.

2. Your task is limited to altering the format while keeping the original meaning and information intact.

3. Please make sure that the revised response can answer the user's question correctly.

4. Please make sure that the revised response is fluent and has no additional subheadings.

5. Evidence is the useful information. You should decide for yourself which parts of the evidence to help rewriting the response.
\\\\
Please first write "Reasoning: $<$reason$>$" to provide a brief reasoning you used to modify, and then write "Revised response: $<$response$>$" to output your final revised response without any additional information, ensuring its fluency.
\\\\
Do not output any additional subheadings.\\
\midrule
\textbf{User Prompt}\\
Below is a user's question, the AI assistant's response, the provided format, and the evidences. 
\\\\

[Question start]

\{question\}

[Question end]

\\\\

[Response start]

\{response\}

[Response end]

\\\\

[Format start]

\{format\}

[Format end]

\\\\

[Evidence start]

\{evidence\}

[Evidence end]
\\
\bottomrule
\end{tabular}
    \caption{Reformatting prompts for knowledge-intensive tasks.}
    \label{tab:rewriting_prompts_retrieval}
\end{table*}

\begin{table*}[t]
    \scriptsize
    \centering
\begin{tabular}{@{}p{1\columnwidth}@{}}
\toprule
\textbf{Query}\\
How many positive three-digit integers less than 500 have at least two digits that are the same?
\\
\midrule
\textbf{Criteria}\\
Task Name: math puzzles

Format: This is a math question. First, analyze the question and give a brief analysis in the first paragraph. Then, use a list to present the step-by-step solution. Next, give another list to output a detailed explanation. Last, give the correct result and a conclusion.\\
\midrule

\textbf{Original Response}\\
To solve this problem, I will count how many three-digit numbers less than 500 have at least two digits that are the same, and subtract that from the total number of three-digit numbers less than 500. The total number of three-digit numbers less than 500 is $4 \times 10 \times 10 = 400$, since the first digit can be 1, 2, 3, or 4, and the second and third digits can be anything from 0 to 9. Now I will count how many of these 400 numbers have no digits that are the same. To have no digits that are the same, the first digit can be any of the 4 choices, but the second digit has to be different from the first, so it has 9 choices. The third digit has to be different from both the first and the second, so it has 8 choices.
\\\\

[* on another note] I am assuming that the second digit can be 0, since the problem does not say otherwise. Therefore, the number of three-digit numbers less than 500 that have no digits that are the same is $4 \times 9 \times 8 = 288$. This means that the number of three-digit numbers less than 500 that have at least two digits that are the same is $400 − 288 = 112$.
\\

\midrule
\textbf{\modelname Response (Ours)}\\
Analysis:

To solve the problem, we need to count how many three-digit numbers less than 500 have at least two digits that are the same.
\\\\
Solution:

1. Calculate the total number of three-digit numbers less than 500: $4 \times 10 \times 10 = 400$.

2. Count the number of three-digit numbers less than 500 that have no digits that are the same: $4 \times 9 \times 8 = 288$.

3. Subtract the count of numbers with no same digits from the total to find the number of three-digit numbers less than 500 that have at least two digits that are the same: $400 - 288 = 112$.
\\\\
Explanation:

- The first digit can be 1, 2, 3, or 4, and the second and third digits can be anything from 0 to 9.

- To have no digits that are the same, the first digit can be any of the 4 choices, but the second digit has to be different from the first, so it has 9 choices. The third digit has to be different from both the first and the second, so it has 8 choices.
\\\\
Result:

There are 112 three-digit numbers less than 500 that have at least two digits that are the same.
\\\\
Conclusion:

There are 112 positive three-digit integers less than 500 that have at least two digits that are the same.
\\
\bottomrule
\end{tabular}
    \caption{A complete example of a query, including the original response and the \modelname response.}
    \label{tab:complete_example}
\end{table*}

\begin{table*}[t]
    \scriptsize
    \centering
\begin{tabular}{@{}p{1\columnwidth}@{}}
\toprule

It is an email-writing task. Here is a general guideline for creating a well-structured and professional email:
\\\\
\ \ \ \ 1. Subject Line: Write a clear and concise subject line that accurately summarizes the content of your email. This helps the recipient understand the purpose of the email at a glance.

\ \ \ \ 2. Salutation: Begin your email with a formal salutation such as "Dear [Recipient's Name]," or use a more casual salutation if you have an informal relationship with the recipient.

\ \ \ \ 3. Introduction: Start your email with a brief introduction, stating who you are and the reason for writing the email. Be clear and to the point, and avoid unnecessary details.

\ \ \ \ 4. Body: This is the main content of your email. Organize your thoughts into paragraphs or bullet points to make them easier to read. Keep your sentences concise and focused. Use proper grammar, punctuation, and spelling to maintain professionalism. If you need to discuss multiple topics, consider using headings or numbered points to separate them.

\ \ \ \ 5. Politeness and Tone: Maintain a polite and respectful tone throughout your email. Be mindful of the recipient's perspective and use appropriate language. Avoid using excessive capitalization, exclamation marks, or emoticons, as they can come across as unprofessional.

\ \ \ \ 6. Closing: Conclude your email with a closing remark, such as "Thank you," or "Best regards," followed by your name. If you expect a response or need specific action, you can mention it in this section as well.

\ \ \ \ 7. Signature: Include your full name, job title, and contact information (e.g., phone number, email address) in your email signature. This helps the recipient easily identify and contact you if needed.

\ \ \ \ 8. Attachments: If you need to include attachments, mention them in the email body and make sure they are relevant to the email's purpose. Consider compressing large files or using cloud storage services if the attachments are too large to be sent via email.

\ \ \ \ 9. Proofread: Before sending the email, proofread it for any grammatical or spelling errors. Make sure the email conveys your message clearly and effectively.
\\\\
The best emails are short, direct, professional, and scannable for the recipient. Follow formal business email structure unless you have an established casual rapport with the recipient.
\\
\bottomrule
\end{tabular}
    \caption{A complete format for ``email generation'' task.}
    \label{tab:complete_format}
\end{table*}

\begin{table*}[t]
    \scriptsize
    \centering
\begin{tabular}{@{}p{1\columnwidth}@{}}
\toprule
\textbf{Query}\\
Is the given statement true or false?

A robotic vacuum can clean both carpets and hard floors.
\\
\midrule
\textbf{Criteria}\\
Task Name: fact verification

Format: This is a fact-verification task. First, give the answer. Then, give an explanation.\\
\midrule
\textbf{Retrieved Evidence}\\
1. Robot vacuums run fully autonomously and automatically. They self-navigate on any flooring type, from hardwoods to carpets, and some models can even be programmed to have a cleaning schedule and to sync to smartphones and smart home devices.

2. iRobot roomba vacuum works on laminate flooring. In fact they work better on hard floor as compared to carpet since their suction and beater bar is not ...

3. Our top pick among the best vacuums for hardwood floors is the iRobot Roomba 694. iRobot is known as a top choice when it comes to robotic ...

4. Because they lack the power to deep clean, though, robot vacuums are best for flooring other than carpet, such as tile, laminate or hardwood, ...\\
\midrule

\textbf{Original Response}\\
True.\\
\midrule
\textbf{\modelname Response (Ours)}\\
True. 

\greenback{Robot vacuums can clean both carpets and hard floors. They are designed to self-navigate on any flooring type, from hardwoods to carpets, and some models can even be programmed to have a cleaning schedule and to sync to smartphones and smart home devices. Additionally, some specific models like the iRobot Roomba 694 are known as top choices for cleaning hardwood floors. However, it is important to note that robot vacuums may lack the power to deep clean carpets compared to hard floors.}
\\
\bottomrule
\end{tabular}
    \caption{An example of a knowledge-intensive task, including the query, the criteria, the retrieved evidence, the original response, and the \modelname response. We highlight the content that is more factual and knowledge in \greenback{green}.}
    \label{tab:kilt_example}
\end{table*}

\begin{table*}[t]
    \scriptsize
    \centering
\begin{tabular}{@{}p{1\columnwidth}@{}}
\toprule
\textbf{System Prompt}\\
Please act as an impartial judge and evaluate the quality of the responses provided by two AI assistants to the user question displayed below. You should choose the assistant that follows the user's instructions and answers the user's question better. Your evaluation should consider the readability including clarity, format, and ease of understanding. Well-organized, grammatically correct response is better. Begin your evaluation by comparing the two responses and provide a short explanation. Avoid any position biases and ensure that the order in which the responses were presented does not influence your decision. Do not allow the length of the responses to influence your evaluation. Do not favor certain names of the assistants. Be as objective as possible. After providing your explanation, output your final verdict by strictly following this format: ``[[A]]'' if assistant A is better, ``[[B]]'' if assistant B is better, and ``[[C]]'' for a tie. \\
\midrule
\textbf{User Prompt}\\

[User Question]

\{question\}

\\\\

[The Start of Assistant A's Answer]

\{answer\_a\}

[The End of Assistant A's Answer]

\\\\

[The Start of Assistant B's Answer]

\{answer\_b\}

[The End of Assistant B's Answer] \\
\bottomrule
\end{tabular}
    \caption{Evaluation prompt for readability.}
    \label{tab:readability_evaluation_prompt}
\end{table*}

\begin{table*}[t]
    \scriptsize
    \centering
\begin{tabular}{@{}p{1\columnwidth}@{}}
\toprule
\textbf{System Prompt}\\
Please act as an impartial judge and evaluate the factuality of the response provided by an AI assistant to the user question displayed below. Your evaluation should consider correctness. You will be given a reference answer and the assistant's answer. Begin your evaluation by comparing the assistant's answer with the reference answer. Identify and correct any mistakes. Be as objective as possible. After providing your explanation, you must rate the response on a scale of 1 to 10 by strictly following this format: ``[[rating]]'', for example: ``Rating: [[5]]''. \\
\midrule
\textbf{User Prompt}\\

[Question]

\{question\}

\\\\

[The Start of Reference Answer]

\{ref\_answer\}

[The End of Reference Answer]

\\\\

[The Start of Assistant's Answer]

\{answer\}

[The End of Assistant's Answer] \\
\bottomrule
\end{tabular}
    \caption{Evaluation prompt for factuality.}
    \label{tab:factuality_evaluation_prompt}
\end{table*}

\begin{table}[t]
    \scriptsize
    \centering
\begin{tabular}{@{}p{1\columnwidth}@{}}
\toprule
\textbf{System Prompt}\\
Please act as a mathematics explanation generator to generate a step-by-step explanation of the answer based on the question presented below.
\\\\
Please follow the instructions below:

1. Please simply generate a step-by-step explanation, including the reason for each step of the calculation.

2. Please do not change the essence of the answer.
\\\\
Please write "The explanation: $<$answer$>$" to output your explanation without any additional information.
\\\\
Here is an example for your reference:
\\\\
Question: Natalia sold clips to 48 of her friends in April, and then she sold half as many clips in May. How many clips did Natalia sell altogether in April and May?
\\\\
Answer: Natalia sold 48/2 = $<<$48/2=24$>>$24 clips in May.$\backslash$nNatalia sold 48+24 = $<<$48+24=72$>>$72 clips altogether in April and May.$\backslash$n\#\#\#\# 72
\\\\
Explanation: 

1. The problem states that Natalia sold clips to 48 friends in April. In May, she sold half as many clips as she did in April. We are asked to find out the total number of clips she sold over these two months.

2. To find out how many clips Natalia sold in May, we take half of the number of clips sold in April. Since she sold clips to 48 friends in April, we calculate half of 48: $\backslash$n\$\$$\backslash$frac\{48\}\{2\} = 24\$\$$\backslash$nThis step involves simple division, where we divide the number of clips sold in April by 2 to find the number of clips sold in May.

3. Now that we have the number of clips sold in May, we need to add this to the number of clips sold in April to get the total number of clips sold over both months:$\backslash$n\$\$48 + 24 = 72\$\$$\backslash$nThis step involves addition, where we sum the clips sold in April and May to find the total sales for the two months combined.

4. Natalia sold 72 clips in total during April and May. This final total gives us the answer to the question asked.\\
\midrule
\textbf{User Prompt}\\
Below is a question and the answer:
\\\\

[Question start]

\{question\}

[Question end]
\\\\

[Answer start]

\{answer\}

[Answer end]\\
\bottomrule
\end{tabular}
    \caption{Prompts for generating explanations for GSM8K.}
    \label{tab:generation_prompts_explanation_gsm8k}
\end{table}

\begin{table}[t]
    \scriptsize
    \centering
\begin{tabular}{@{}p{1\columnwidth}@{}}
\toprule
\textbf{System Prompt}\\
Please act as a mathematics explanation generator to generate a step-by-step explanation of the answer based on the question presented below.
\\\\
Please follow the instructions below:

1. Please simply generate a step-by-step explanation, including the reason for each step of the calculation.

2. Please do not change the essence of the answer.
\\\\
Please write "The explanation: $<$answer$>$" to output your explanation without any additional information.
\\\\
Here is an example for your reference:
\\\\
Question: Each row of a seating arrangement seats 7 or 8 people. Forty-six people are to be seated. How many rows seat exactly 8 people if every seat is occupied?
\\\\
Answer: Let \$x\$ be the number of rows with 8 people.  If we removed a person from each of these rows, then every row would contain 7 people.  Therefore, \$46 - x\$ must be divisible by 7.$\backslash$n$\backslash$nThen \$x $\backslash$equiv 46 $\backslash$equiv 4 $\backslash$pmod\{7\}\$.  The first few positive integers that satisfy this congruence are 4, 11, 18, and so on.  However, each row contains at least 7 people.  If there were 7 or more rows, then there would be at least \$7 $\backslash$cdot 7 = 49\$ people.  We only have 46 people, so there must be at most six rows.  Therefore, the number of rows with 8 people is \$$\backslash$boxed\{4\}\$.$\backslash$n\#\#\#\# 4
\\\\
Explanation: 

We have a seating arrangement where each row seats either 7 or 8 people. A total of 46 people need to be seated, and all seats must be filled. The question asks how many rows seat exactly 8 people.

1. We let \$x\$ represent the number of rows that seat exactly 8 people. 

2. To simplify the problem, imagine removing one person from each row that currently seats 8 people. This would convert every 8-person row into a 7-person row. Now, all rows (both the original 7-person rows and the adjusted 8-person rows) would seat 7 people.

3. With this adjustment, the total number of people would be reduced by \$x\$ (since we removed one person from each 8-person row), making it \$46 - x\$.

4. Since all rows now hypothetically seat 7 people, the adjusted total, \$46 - x\$, must be divisible by 7 for it to be a possible total number of people seated in rows of 7.

5. We analyze \$46 - x\$ in terms of modulo 7. Specifically, we want to find values of \$x\$ such that \$46 - x\$ is a multiple of 7. This simplifies to finding \$x\$ such that \$x $\backslash$equiv 46 $\backslash$pmod\{7\}\$. Calculating \$46 $\backslash$mod 7\$ yields 4, because when 46 is divided by 7, the remainder is 4. This tells us \$x\$ must be equivalent to 4 modulo 7.

6. The numbers that satisfy \$x $\backslash$equiv 4 $\backslash$pmod\{7\}\$ are 4, 11, 18, etc. However, we need a practical value of \$x\$ that fits the total people and the row seating constraint.

7. If there were 7 or more rows of 7 people each, we'd have at least 49 people (since \$7 $\backslash$times 7 = 49\$). But we only have 46 people, so there must be fewer than 7 rows in total. The plausible values of \$x\$ from the list 4, 11, 18, etc., must be reconsidered within this context.

8. Since the only value from our possible \$x\$ values (4, 11, 18) that is less than 7 and fits the total people count is 4, we conclude that there are 4 rows of 8 people each.
\\
\midrule
\textbf{User Prompt}\\
Below is a question and the answer:
\\\\

[Question start]

\{question\}

[Question end]
\\\\

[Answer start]

\{answer\}

[Answer end]\\
\bottomrule
\end{tabular}
    \caption{Prompts for generating explanations for MATH.}
    \label{tab:generation_prompts_explanation_math}
\end{table}

\end{document}